\documentclass[sigconf, screen, nonacm]{acmart}
\renewcommand\footnotetextcopyrightpermission[1]{}
\usepackage{booktabs}   % 提供三线表指令 (\toprule, \midrule 等)
\usepackage{multirow}   % 跨行合并（虽然图里主要是跨列）
\usepackage[table]{xcolor} % 提供颜色支持 (\rowcolor)
\definecolor{grey}{rgb}{0.5,0.5,0.5} % define grey color
\usepackage{pifont}
\usepackage{graphicx} % 必须引用，用于缩放表格内容
\usepackage{makecell} % 用于处理复杂的单元格
\usepackage{listings}
\usepackage[skins,breakable,most]{tcolorbox}
\usepackage[ruled,vlined]{algorithm2e}
\usepackage{svg} % For \includesvg command
\usepackage{adjustbox}
\usepackage{caption}  % 在导言区添加，如果还没有的话
\usepackage{titletoc}
\usepackage[title]{appendix}

\AtBeginDocument{%
  }

\acmSubmissionID{2868}

\begin{document}

\title{AnomalyAgent: Agentic Industrial Anomaly Synthesis via Tool-Augmented Reinforcement Learning}

\author{Jiaming Su}
\affiliation{%
  \institution{Fudan University}
  \city{Shanghai}
  \country{China}
}
\email{jiamingsu08@gmail.com}
\orcid{0009-0002-4813-4051}

\author{Tengchao Yang}
\affiliation{%
  \institution{Tongji University}
  \city{Shanghai}
  \country{China}
}
\email{krisy0409@gmail.com}

\author{Ruikang Zhang}
\affiliation{%
  \institution{Peking University}
  \city{Beijing}
  \country{China}
}
\email{2601213508@stu.pku.edu.cn}

\author{Zhengan Yan}
\affiliation{%
  \institution{Shanghai Jiaotong University}
  \city{Shanghai}
  \country{China}
}
\email{zhenganyan@sjtu.edu.cn}

\author{Haoyu Sun}
\affiliation{%
  \institution{Fudan University}
  \city{Shanghai}
  \country{China}
}
\email{25211010070@m.fudan.edu.cn}

\author{Linfeng Zhang}
\correspondingauthor
\affiliation{
  \institution{Shanghai Jiaotong University}
  \city{Shanghai}
  \country{China}
}
\email{zhanglinfeng1997@outlook.com}
\renewcommand{\shortauthors}{Jiaming Su et al.}

\begin{abstract}
Industrial anomaly generation is a crucial method for alleviating the data scarcity problem in anomaly detection tasks. Most existing anomaly synthesis methods rely on single-step generation mechanisms, lacking complex reasoning and iterative optimization capabilities, making it difficult to generate anomaly samples with high semantic realism. We propose AnomalyAgent, an anomaly synthesis agent with self-reflection, knowledge retrieval, and iterative refinement capabilities, aiming to generate realistic and diverse anomalies. Specifically, AnomalyAgent is equipped with five tools: Prompt Generation (PG), Image Generation (IG), Quality Evaluation (QE), Knowledge Retrieval (KR), and Mask Generation (MG), enabling closed-loop optimization. To improve decision-making and self-reflection, we construct structured trajectories from real anomaly images and design a two-stage training framework: supervised fine-tuning followed by reinforcement learning. This process is driven by a three-part reward mechanism: (1) task rewards to supervise the quality and location rationality of generated anomalies; (2) reflection rewards to train the model's ability to improve anomaly synthesis prompt; (3) behavioral rewards to ensure adherence to the trajectory. On the MVTec-AD dataset, AnomalyAgent achieves IS/IC-L of 2.10/0.33 for anomaly generation, 57.0\% classification accuracy using ResNet34, and 99.3\%/74.2\% AP at the image/pixel level using a simple UNet, surpassing all zero-shot SOTA methods. The code is publicly available at \href{https://github.com/sssjmmm/AnomalyAgent}{https://github.com/sssjmmm/AnomalyAgent}.
\end{abstract}





\maketitle

\begin{figure}[t]
    \centering
    \includegraphics[width=\linewidth]{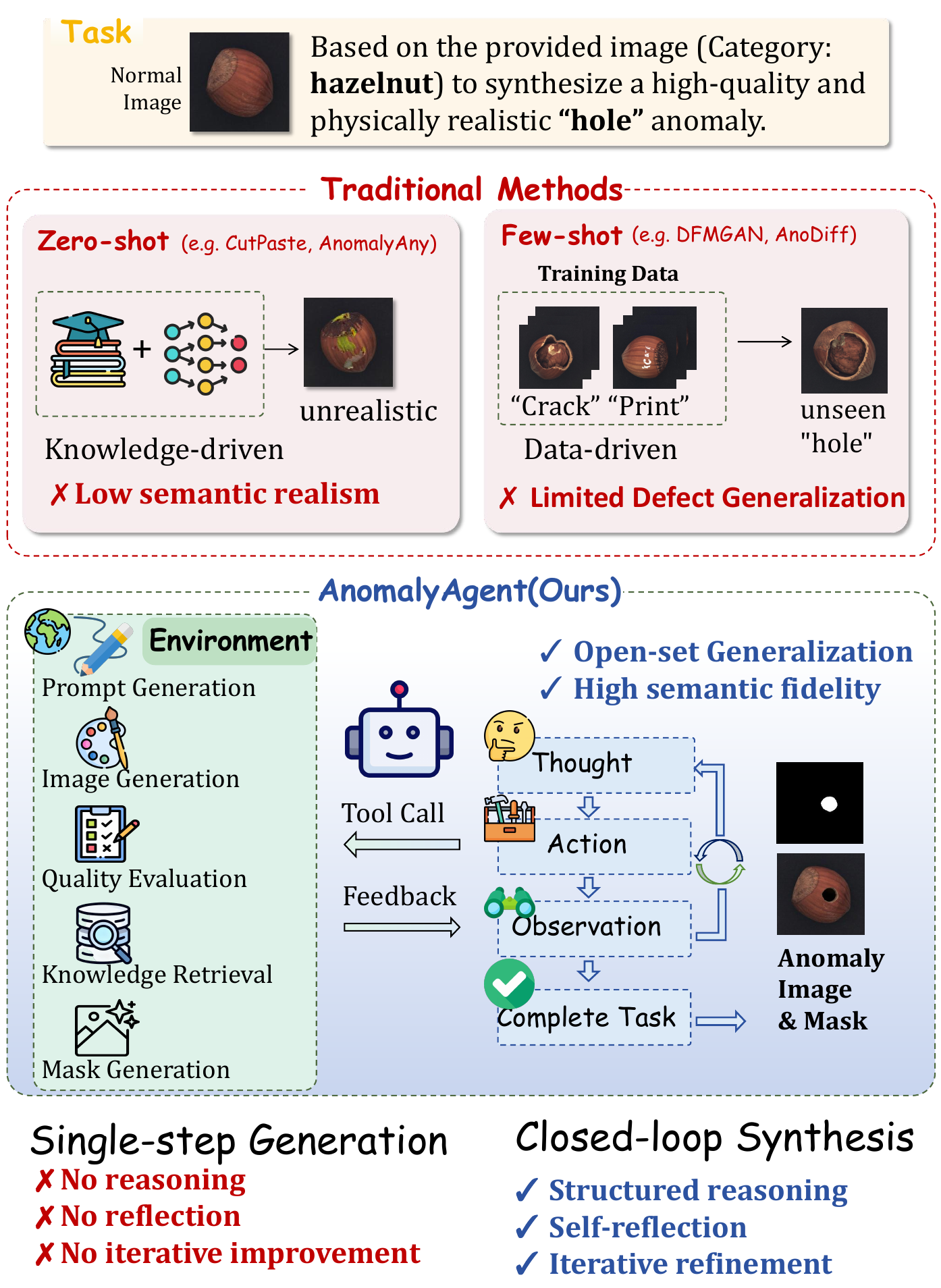}
    \caption{Motivation of AnomalyAgent.
Few-shot methods generalize poorly with limited defect data, while zero-shot methods often generate semantically inconsistent anomalies. AnomalyAgent resolves this via a closed-loop agentic framework for realistic, generalizable synthesis.}
    \vspace{-3pt}
    \label{fig:motivation}
\end{figure}

\section{Introduction}

\begin{figure*}[t]
  \centering
  \includegraphics[width=\linewidth]{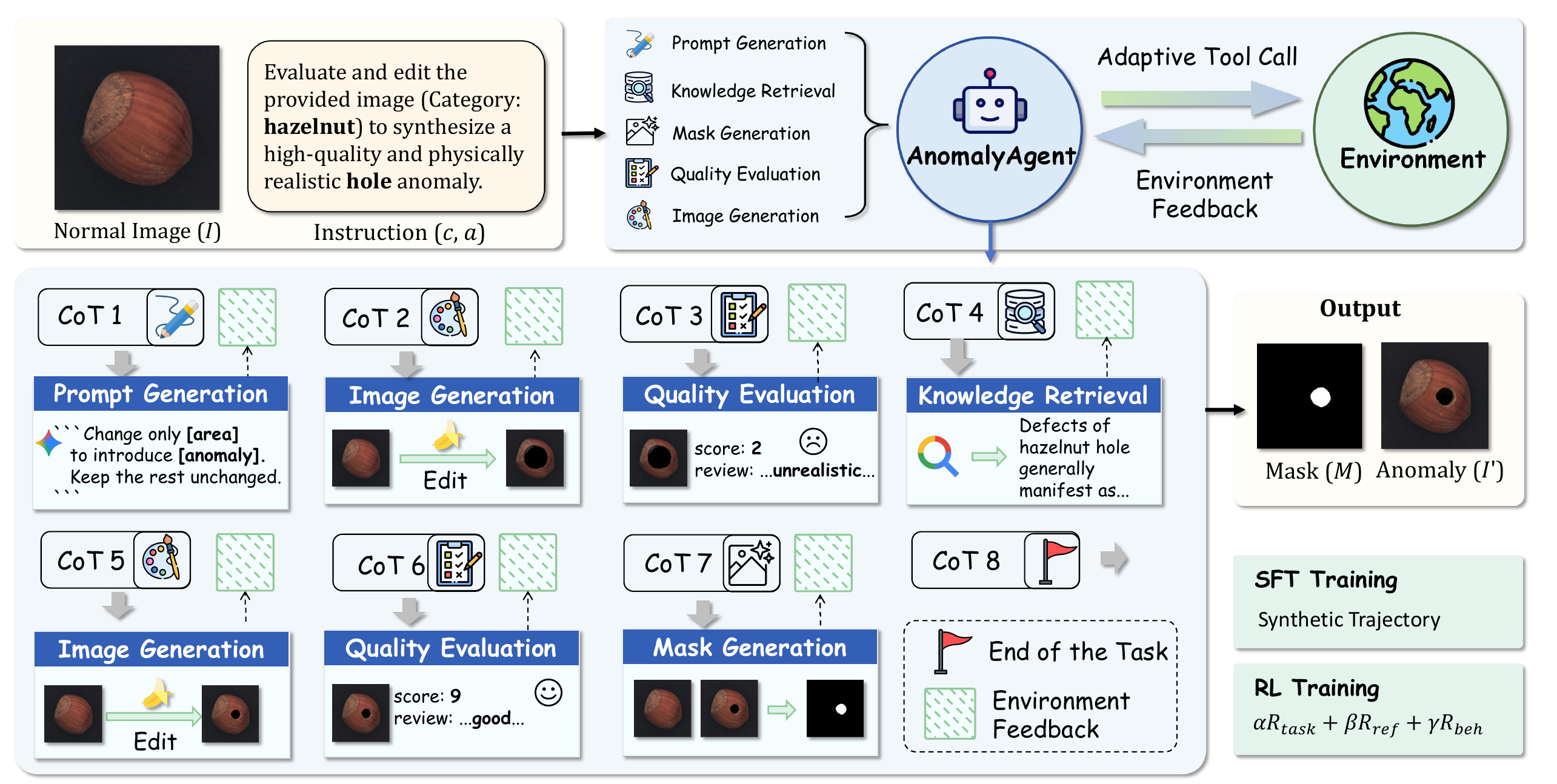}
  \caption{Overview of AnomalyAgent. Given a normal image, the agent iteratively invokes tools (PG, IG, QE, KR, MG) through a thought–action–observation loop. Feedback from each turn guides refinement, producing high-quality anomaly images and corresponding masks. \underline{Tool selection is adaptively determined by the model instead of a predefined sequence.}}
  \Description{Overview of AnomalyAgent. 
  A normal industrial image and an anomaly prompt are input to the agent planner,
  which selects appropriate tools from a tool library to synthesize anomaly images
  and corresponding masks.}
  \label{fig:overview}
\end{figure*}

Industrial Anomaly Detection (IAD) plays a pivotal role in intelligent manufacturing. Despite recent progress\cite{DBLP:journals/corr/abs-2502-16412}, real-world industrial scenarios are characterized by extreme scarcity and high diversity of anomalous samples, leading to severely imbalanced data distributions. This inherent limitation significantly hinders the generalization capability of anomaly detection models. To alleviate this problem, anomaly synthesis has become a current research hotspot, aiming to provide crucial supervisory signals to downstream models by generating high-fidelity simulated defects.

As illustrated in Fig. \ref{fig:motivation}, existing anomaly synthesis methods can be broadly categorized into few-shot and zero-shot paradigms depending on whether anomaly samples are available. Few-shot approaches\cite{DBLP:conf/wacv/ZhangCHL21, DBLP:conf/aaai/DuanH0023, DBLP:conf/aaai/HuZYDCLWW24, DBLP:conf/eccv/YangCCFLLC24, DBLP:conf/eccv/GuiGLWW24, DBLP:conf/cvpr/SongPB0C0Y25, DBLP:journals/corr/abs-2410-14987, DBLP:conf/cvpr/JinPHHWCWZCLW25, DBLP:conf/icmcs/LinCZL21} mainly utilize GANs\cite{DBLP:journals/corr/GoodfellowPMXWOCB14} or diffusion models\cite{DBLP:conf/nips/HoJA20} to capture specific defect distributions, but their generation capability is limited by the types of anomalies observed during training. In contrast, zero-shot methods\cite{DBLP:conf/cvpr/LiSYP21, DBLP:conf/iccv/ZavrtanikKS21, DBLP:conf/eccv/SchluterTHK22, DBLP:conf/cvpr/Zhang0Z24, DBLP:conf/cvpr/SunCDF25, DBLP:conf/cvpr/Zhao25} operate under more stringent real-world constraints, constructing anomalies by heuristically perturbing normal images (e.g., CutPaste\cite{DBLP:conf/cvpr/LiSYP21}, DRAEM \cite{DBLP:conf/iccv/ZavrtanikKS21}) or utilizing multimodal generation priors (e.g., AnomalyAny\cite{DBLP:conf/cvpr/SunCDF25}). Nevertheless, zero-shot approaches often suffer from a lack of semantic realism. Despite some progress, most methods, whether zero-shot or few-shot, adopt a single-step generation paradigm, where anomalies are synthesized in a one-pass manner without subsequent feedback or refinement. Such a process can be characterized as an \emph{open-loop} generation scheme. In this scenario, the models are unable to perform self-reflection, knowledge retrieval, and iterative refinement. Therefore, the generation process is difficult to control, often resulting in unrealistic structural and semantic inconsistencies. The limitations of this \emph{open-loop} approach prevent generative models from fully realizing their expressive potential.

Recently, the emergence of Agentic Reinforcement Learning has introduced a promising paradigm to address the challenges above. Reasoning models, exemplified by OpenAI o1\cite{DBLP:journals/corr/abs-2412-16720} and DeepSeek-R1\cite{DBLP:journals/corr/abs-2501-12948}, have demonstrated that optimizing long-horizon decision trajectories via reinforcement learning can substantially enhance a model’s capability for complex logical reasoning. However, in the field of industrial anomaly synthesis, how to deeply couple planning and reasoning capabilities with image editing tools to construct an autonomous synthetic agent with a ``perception-reflection-action'' closed loop remains a challenge that has not yet been fully explored. Existing generative frameworks often face capability fragmentation: Multimodal Large Language Models (MLLMs) are good at planning but struggle to directly generate high-fidelity images, while general image generation and editing models possess powerful image generation capabilities but lack reasoning abilities, heavily relying on the fine-grained control of high-quality prompts.

To bridge this capability gap, we propose \textbf{AnomalyAgent (Industrial Anomaly Synthesis Agent)}, the first tool-driven multimodal agent specifically designed for industrial anomaly synthesis. AnomalyAgent reformulates anomaly synthesis as a sequential decision-making task, equipping the MLLM with five tools: Prompt Generation (PG), Image Generation (IG), Quality Evaluation (QE), Knowledge Retrieval (KR), and Mask Generation (MG). This design enables the agent to dynamically make decisions and perform autonomous self-reflection throughout the process. Through this tool-integrated reasoning approach, AnomalyAgent can proactively guide generative model and achieve iterative optimization through a sequence of planning, generation, self-reflection, and improvement within a closed-loop framework. To effectively train the proposed agent, we first design a trajectory construction strategy that transforms real anomaly samples into structured multi-turn reasoning trajectories. Specifically, we synthesize normal images reversely based on real anomaly images, and reconstruct the anomaly generation process through an N-step paradigm to obtain trajectories without requiring additional manual annotations. Building upon this, we adopt a two-stage training pipeline. The first stage involves cold-start supervised fine-tuning (SFT) on the constructed trajectories, enabling the model to master the format and basic strategies of tool invocations. The second stage introduces Reinforcement Learning with Verifiable Rewards (RLVR)~\cite{DBLP:journals/corr/abs-2402-03300}, which optimizes the agent's decision-making strategy and self-reflection mechanism through trajectory-level reinforcement learning. This allows the agent to adapt its generation strategy based on environmental feedback, significantly improving the realism and diversity of synthesized samples and enhancing downstream task performance.

AnomalyAgent achieves performance exceeding previous zero-shot methods on the MVTec-AD\cite{DBLP:journals/ijcv/BergmannBFSS21} benchmark, demonstrating that Agentic RL-driven anomaly synthesis can generate more challenging and informative supervisory signals for downstream tasks, and offers a new paradigm for zero-shot industrial anomaly detection.

Our main contributions are summarized as follows:

\begin{itemize}
    \item We propose \textbf{AnomalyAgent}, the first agentic framework for industrial anomaly synthesis, which formulates synthesis as a sequential decision-making process with coordinated multimodal tools.

    \item We introduce a trajectory construction strategy based on real anomaly images, enabling scalable training without additional manual annotation.

    \item Our method surpasses the zero-shot SOTA across 15 object categories. AnomalyAgent outperforms \emph{Gemini 3.1 Flash Image Preview} by 12.3\% in downstream classification and achieves a 4.2\% higher image-level AUC than conventional anomaly detection baselines.
\end{itemize}

\raggedbottom
\section{Related Work}

\subsection{Agentic Reinforcement Learning}

Agentic Reinforcement Learning (Agentic RL) \cite{DBLP:journals/tmlr/ZhangGYYZTZLXLZCZFWHVLW26, DBLP:journals/corr/abs-2510-04206} marks a paradigm shift from passive modeling to autonomous decision-making by optimizing interactive trajectories involving multi-turn reasoning, tool use, and environmental feedback. Early paradigms such as ReAct \cite{DBLP:conf/iclr/YaoZYDSN023} and Chain-of-Thought (CoT) \cite{DBLP:conf/nips/Wei0SBIXCLZ22} establish the \textit{Thought-Action-Observation} loop, while SFT-based methods \cite{DBLP:conf/iclr/QinLYZYLLCTQZHT24} suffer from limited generalization and poor error recovery. Recent advances in reinforcement fine-tuning (RFT) and Tool-Integrated Reasoning (TIR) (e.g. SimpleTIR \cite{DBLP:journals/corr/abs-2509-02479}, AutoTIR\cite{DBLP:journals/corr/abs-2507-21836}) demonstrate that reinforcement signals enable autonomous strategy discovery beyond fixed heuristics. In multimodal settings, models such as Pixel-Reasoner\cite{DBLP:journals/corr/abs-2505-15966} and DeepEyes \cite{DBLP:journals/corr/abs-2505-14362} align visual perception with reasoning and tool execution. Optimizing long-horizon trajectories remains challenging due to credit assignment. While proximal policy optimization (PPO) \cite{DBLP:journals/corr/SchulmanWDRK17} is widely used,  Group Relative Policy Optimization (GRPO) \cite{DBLP:journals/corr/abs-2402-03300} improves stability via group-wise normalization and has been adopted in reasoning models such as OpenAI o1 \cite{DBLP:journals/corr/abs-2412-16720}, DeepSeek-R1 \cite{DBLP:journals/corr/abs-2501-12948}, and Vision-R1\cite{DBLP:journals/corr/abs-2503-06749}. Extensions with memory and reflection, including DeepEyesV2 \cite{DBLP:journals/corr/abs-2511-05271}, Memory-R1\cite{DBLP:journals/corr/abs-2508-19828}, MEM1\cite{DBLP:journals/corr/abs-2506-15841}, Voyager \cite{DBLP:journals/tmlr/WangX0MXZFA24}, and Reflexion \cite{DBLP:conf/nips/ShinnCGNY23}, further enhance long-horizon decision-making through iterative perception-action cycles. To address sparse rewards, recent works introduce semantic supervision such as LLM-as-a-Judge \cite{DBLP:conf/nips/ZhengC00WZL0LXZ23}, self-correction \cite{DBLP:journals/corr/abs-2308-03188}, and reflection-based rewards \cite{DBLP:conf/nips/ShinnCGNY23}.

\begin{algorithm}[t]
\caption{AnomalyAgent Trajectory}
\label{alg:agent}
\KwIn{Normal image $I$, category $c$, anomaly type $a$}
\KwOut{Anomaly image $I'$, mask $M$}

Initialize prompt: $p_0 \leftarrow \text{PG}(I, c, a)$

\For{$t = 1$ \KwTo $N$}{
    $I'_t \leftarrow \text{IG}(I, p_{t-1})$ \\
    $(s_t, f_t) \leftarrow \text{QE}(I, I'_t)$ \tcp*{score \& feedback}
    
    \If{$s_t \ge \theta$}{
        \textbf{break}
    }
    
    \If{NeedKR$(f_t)$}{
        $k \leftarrow \text{KR}(c, a)$ \tcp*{optional KR}
        $p_t \leftarrow \text{Refine}(p_{t-1}, f_t, k)$
    }
    \Else{
        $p_t \leftarrow \text{Refine}(p_{t-1}, f_t)$
    }
}

$M \leftarrow \text{MG}(I, I'_t)$

\Return{$(I'_t, M)$}

\end{algorithm} 
\vspace{-10pt}

\subsection{Industrial Anomaly Synthesis}

Industrial Anomaly Synthesis (IAS) mitigates the scarcity of defect data by generating synthetic anomalies. Existing approaches can be divided into few-shot and zero-shot paradigms.

\textbf{Few-shot methods} model the distribution of limited anomalous samples. Early works rely on GANs \cite{DBLP:journals/corr/GoodfellowPMXWOCB14}, such as Defect-GAN \cite{DBLP:conf/wacv/ZhangCHL21} and DFMGAN \cite{DBLP:conf/aaai/DuanH0023}. More recent approaches adopt diffusion models \cite{DBLP:conf/nips/HoJA20}, including AnoDiff \cite{DBLP:conf/aaai/HuZYDCLWW24} and Defect-Gen \cite{DBLP:conf/eccv/YangCCFLLC24}, or reformulate synthesis as inpainting \cite{DBLP:conf/eccv/GuiGLWW24, DBLP:conf/cvpr/SongPB0C0Y25}. Extensions such as DualAnoDiff \cite{DBLP:conf/cvpr/JinPHHWCWZCLW25} and SeaS \cite{DBLP:journals/corr/abs-2410-14987} improve controllability and disentanglement. However, their performance is fundamentally limited by the diversity of available defect samples.

\textbf{Zero-shot methods} eliminate the need for anomalous data. Heuristic approaches such as CutPaste \cite{DBLP:conf/cvpr/LiSYP21}, NSA \cite{DBLP:conf/eccv/SchluterTHK22}, and DRAEM \cite{DBLP:conf/iccv/ZavrtanikKS21} generate pseudo anomalies via handcrafted perturbations, but often lack semantic realism. Recent generative approaches leverage pretrained priors: RealNet \cite{DBLP:conf/cvpr/Zhang0Z24} perturbs diffusion trajectories, while AnomalyAny \cite{DBLP:conf/cvpr/SunCDF25} utilizes Stable Diffusion \cite{DBLP:conf/cvpr/RombachBLEO22} for prompt-driven synthesis. AnoStyler ~\cite{DBLP:journals/corr/abs-2511-06687} frames anomaly generation as text-guided localized style transfer. AnoHybrid \cite{DBLP:conf/cvpr/Zhao25} further combines heuristic perturbations with generative priors to balance controllability and realism, yet still rely on static pipelines without iterative refinement.

Despite these advances, most existing methods rely on static, single-step generation without feedback or refinement, limiting controllability and realism in complex industrial scenarios. To address this, we propose AnomalyAgent, which formulates anomaly synthesis as an agentic decision-making process with iterative planning, tool use, and feedback refinement.

\section{Method}

\subsection{Overview of AnomalyAgent}

Industrial anomaly synthesis requires both visual realism and semantic consistency, yet conventional single-step pipelines often produce unrealistic or mismatched defects. We propose \textbf{AnomalyAgent}, which formulates synthesis as a multi-turn, tool-guided reasoning process to enable iterative refinement.
\begin{figure*}
    \centering
    \includegraphics[width=0.91\linewidth]{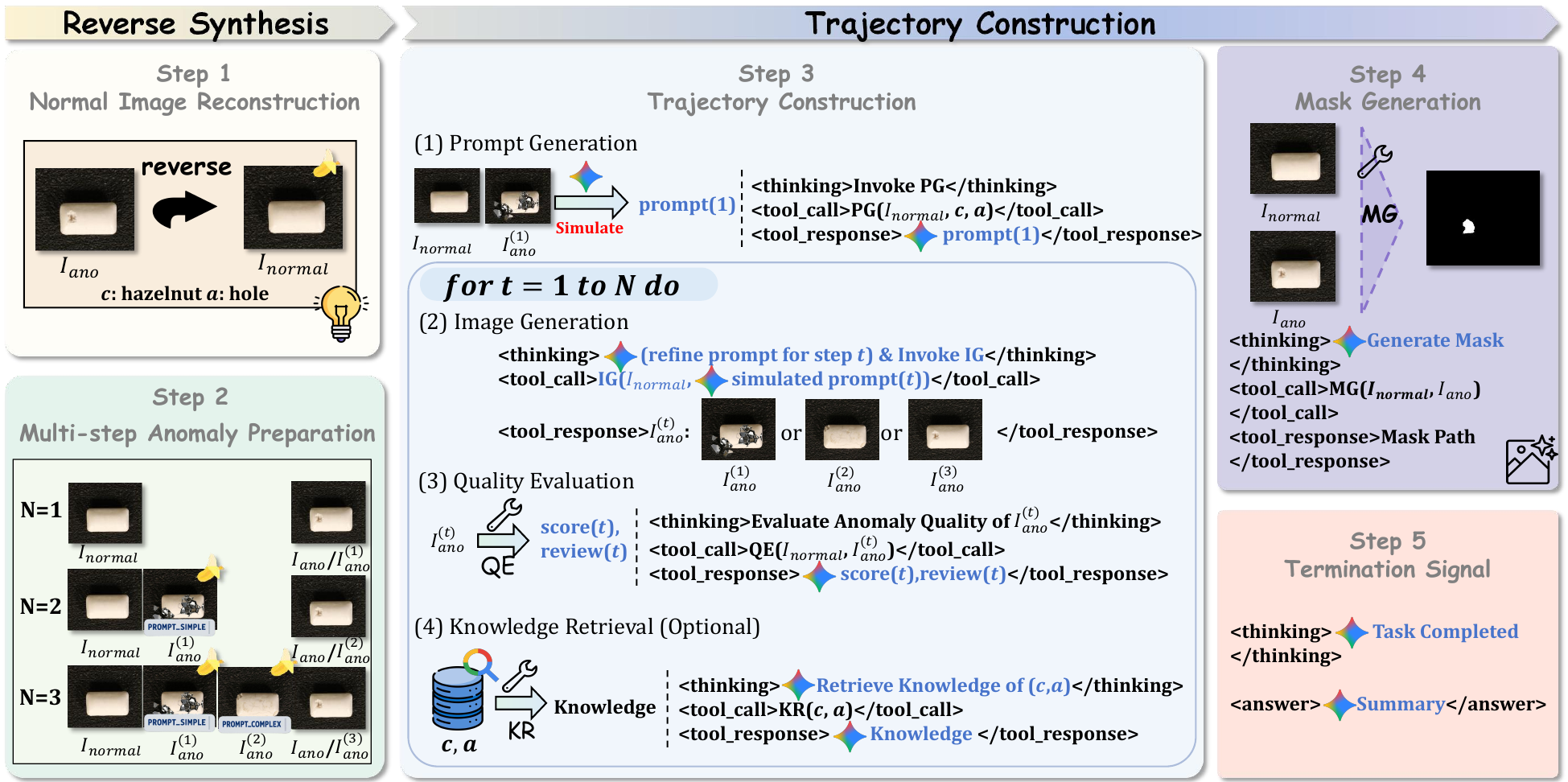}
    \caption{Pipeline of trajectory construction and taxonomy. Given an anomaly image, we generate multi-turn trajectories categorized into three types based on the number of IG calls.}
    \label{fig:sft trajectory}
\end{figure*}

As illustrated in Fig.~\ref{fig:overview}, given a normal industrial image $I$, the corresponding object category $c$, and a target anomaly type $a$, AnomalyAgent dynamically coordinates multiple tools to produce a high-quality anomaly image $I'$ along with a corresponding anomaly mask $M$. The framework integrates five specialized tools:
\begin{itemize}
\item \textbf{Prompt Generation (PG).}
This tool generates an initial textual prompt describing the desired anomaly. Given $(I, c, a)$, a large language model produces a structured prompt $p$ that captures the expected visual characteristics of the defect.

\item \textbf{Image Generation (IG).}
Conditioned on the original image $I$ and the generated prompt $p$, an image generation and editing model synthesizes an anomaly image $I'$. This module injects defects while preserving the global structure and appearance of the original object.

\item \textbf{Quality Evaluation (QE).}
AnomalyAgent employs an evaluation module that analyzes both the original image $I$ and the generated image $I'$. The evaluator outputs a quality score $s$ together with textual feedback $f$ indicating alignment with the target anomaly type and visual plausibility.

\item \textbf{Knowledge Retrieval (KR).}
This module retrieves textual descriptions of typical defects given $(c, a)$ from a knowledge base, providing semantic guidance for prompt refinement and ensuring cross-modal alignment.

\item \textbf{Mask Generation (MG).}
AnomalyAgent generates a pixel-level anomaly mask $M$ by comparing the original image $I$ with the synthesized anomaly $I'$. The mask accurately localizes the injected defect to support downstream tasks.
\end{itemize}

AnomalyAgent performs iterative reasoning and multi-tool interactions, which can be naturally formalized as a sequential decision process. The overall procedure is summarized in Algorithm~\ref{alg:agent}. Upon initial Prompt Generation, the agent synthesizes a candidate anomaly image and invokes Quality Evaluation to obtain a numerical quality score along with semantic feedback. If the score falls below a predefined threshold, the agent enters a self-correction phase, where it assesses the discrepancy between the current prompt and the generated anomalies, and iteratively refines the prompt. This refinement is driven by the internal reasoning of agent and optionally by knowledge retrieval to acquire complementary domain knowledge, enabling more informed and targeted prompt updates. This perception-reflection-action loop continues until the synthesized anomaly satisfies quality constraints or the maximum number of iterations is reached. Finally, the agent invokes Mask Generation to produce a pixel-level anomaly mask for the accepted image, ensuring precise structural and semantic alignment.

Training proceeds in two complementary stages. First, Supervised Fine-Tuning (SFT) equips the model with structured tool-use patterns by leveraging synthesis trajectories constructed from real anomaly data. Second, Agentic Reinforcement Learning (RL) further refines long-horizon decision-making through environment interaction. In the RL stage, a multi-component reward supervises three critical aspects: \textbf{task reward} encourages high-quality anomaly generation, \textbf{reflection reward} promotes iterative prompt refinement, and \textbf{behavioral reward} ensures disciplined adherence to tool invocation sequences.

\subsection{Supervised Fine-Tuning Stage}

\textbf{Trajectory Construction and Taxonomy.}
We construct multi-turn synthesis trajectories for supervised training based on real anomalous samples from the VisA\cite{DBLP:conf/eccv/ZouJPZD22} dataset. Specifically, starting from each anomaly image $I_{\text{ano}}$, we use a image generation and editing model for reverse image generation and MLLM for textual generation to synthesize a complete reasoning trajectory, including CoT, tool invocation arguments, and corresponding tool outputs. Each trajectory is organized in a structured format, where the reasoning process is encapsulated within \texttt{<thinking>...</thinking>} and the tool invocation is represented as \texttt{<tool\_call>...</tool\_call>}, enabling explicit modeling of decision-making and tool interaction.

As illustrated in Fig. \ref{fig:sft trajectory}, for each $I_{\text{ano}}$, we first obtain a corresponding normal image $I_{\text{normal}}$ via reverse synthesis, which serves as a clean reference to facilitate high-quality prompt generation aligned with real data distribution. Building upon this, we employ a unified \emph{N-step generation} paradigm to generate trajectories of varying complexity. For a trajectory with $N$ Image Generation turns, we denote the intermediate anomaly image at turn $t$ as $I_{\text{ano}}^{(t)}$ ($t = 1, \dots, N$), where $I_{\text{ano}}^{(1)}$ is generated from $I_{\text{normal}}$ using a simple prompt, and subsequent images are obtained through iterative prompt refinement. The final target $I_{\text{ano}}^{(N)}$ is given as the ground-truth anomaly image $I_{\text{ano}}$. Given the image sequence ${I_{\text{normal}}, I_{\text{ano}}^{(1)}, \dots, I_{\text{ano}}^{(N)}}$, we construct trajectories in a forward manner by sequentially generating CoT, tool invocation arguments, and tool outputs. Specifically, both the prompt generated by Prompt Generation and the prompt refined by the agent through self-reflection are derived from currently available anomaly images, thereby enforcing alignment between prompt descriptions and visual defect patterns. Meanwhile, Quality Evaluation provides scoring feedback based on the anomaly images generated in the current turn, Knowledge Retrieval supplies auxiliary semantic context when needed, and Mask Generation produces anomaly masks at the final stage via hard coding. Finally, an answer is generated by summarizing the entire trajectory, yielding a coherent description that integrates intermediate reasoning turns and tool interaction outcomes. The environment monitors the generated output and terminates the interaction once the special token \texttt{<answer>...</answer>} is detected. For details regarding the pseudocode, please refer to the attached materials.

We categorize trajectories by the number of IG calls into three types with increasing synthesis difficulty, offering a staged guidance from simple generation to reflective optimization. This hierarchical design establishes the foundation for reinforcement learning while maintaining this progressive guidance:
% \vspace{-20pt}

\begin{itemize}
    \item \textbf{Single-Generation}: Only one IG is needed to generate the required result, following PG $\rightarrow$ IG $\rightarrow$ QE $\rightarrow$ MG.
    \item \textbf{Dual-Generation}: the first attempt fails QE, prompting refinement (optionally with KR), following PG $\rightarrow$ IG $\rightarrow$ QE $\rightarrow$ (KR) $\rightarrow$ IG $\rightarrow$ QE $\rightarrow$ MG.
    \item \textbf{Triple-Generation}: more challenging cases requiring multiple refinements, following PG $\rightarrow$ IG $\rightarrow$ QE $\rightarrow$ KR $\rightarrow$ IG $\rightarrow$ QE $\rightarrow$ IG $\rightarrow$ QE $\rightarrow$ MG.
\end{itemize}

\noindent\textbf{Supervised Fine-Tuning.}
After constructing reasoning trajectories, we perform cold-start SFT to initialize the agent with stable tool-use behavior. The model is trained to autoregressively predict the next action, including both textual reasoning turns and structured tool invocations, conditioned on the current context and input image. Formally, given an image $x$ and a trajectory $\tau=\{(s_t,a_t)\}_{t=1}^T$, the objective is:
\begin{equation}
\mathcal{L}_{\text{SFT}} = - \sum_{t=1}^{T} \log p_{\theta}(a_t \mid s_t, x),
\end{equation}
where $s_t$ denotes the current state, and this stage learns to imitate expert trajectories, providing a strong initialization for RL.

\subsection{Agentic Reinforcement Learning}

While SFT enables the model to imitate expert trajectories, it remains limited to passive pattern replication. To optimize decision quality, we introduce an agentic reinforcement learning phase where the agent interacts with the environment through iterative reasoning and tool usage. In each turn, the agent dynamically invokes various tools to progressively improve anomaly synthesis.

\noindent\textbf{RL Algorithm}.
To optimize tool-use sequences, we employ GRPO\cite{DBLP:journals/corr/abs-2402-03300}, which eliminates the Critic network by utilizing relative feedback within a group of $G$ independent trajectories $\{\tau_i\}_{i=1}^G$. The advantage $A_i$ is estimated by normalizing rewards within the group: $A_i = [r(\tau_i) - \mathrm{mean}(r)] / \mathrm{std}(r)$. The objective maximizes a clipped surrogate loss while constraining the policy $\pi_\theta$ within the proximity of a reference policy $\pi_{\text{ref}}$:
\begin{equation}
\begin{aligned}
\mathcal{L}_{\text{GRPO}} = \mathbb{E}_{\tau_i \sim \pi_{\text{old}}, t} \Bigl[ & \min \left( \rho_{i,t} A_i, \text{clip}(\rho_{i,t}, 1-\epsilon, 1+\epsilon) A_i \right) \\
& - \beta D_{KL}(\pi_\theta(\cdot|x_t) \parallel \pi_{\text{ref}}(\cdot|x_t)) \Bigr],
\end{aligned}
\end{equation}
where $\rho_{i,t} = \frac{\pi_\theta(a_t|x_t)}{\pi_{\text{old}}(a_t|x_t)}$ denotes the importance sampling ratio. This group-based mechanism enables AnomalyAgent to efficiently explore the tool-action space and identify optimal synthesis paths with minimal memory overhead.

\noindent\textbf{Reward Design.}
Effective anomaly synthesis requires the agent to solve three fundamental questions: \emph{Is the generated anomaly authentic?} \emph{Can it be improved?} \emph{And is the generation process well executed?} These correspond to three key capabilities: quality assessment, iterative improvement, and disciplined decision-making. Accordingly, we design a unified reward function that explicitly models these aspects, enabling the agent to learn not only what to generate, but also how to refine and when to act:
\begin{equation}
R = \alpha R_{\text{task}} + \beta R_{\text{ref}} + \gamma R_{\text{beh}},
\end{equation}
where $\alpha$, $\beta$, and $\gamma$ are weighting coefficients.

\textbf{Task Reward.}
We evaluate the realism and industrial plausibility of the final generated anomaly using \emph{LLM-as-a-Judge}:
\begin{equation}
R_{\text{task}} = S_{\text{final}},
\end{equation}
where $S_{\text{final}}$ is a scalar score measuring both defect localization rationality and visual fidelity of the final output image.

\textbf{Reflection Reward.}
Let $m$ denote the index of images generated within a single trajectory. To encourage iterative refinement, we reward improvements across consecutive generation turns:
\begin{equation}
    R_{\text{ref}} = \sum_{m=1}^{M} \max(0, S_{m} - S_{m-1}),
\end{equation}
where $S_m$ denotes the quality score of the $m$-th synthesized anomaly. This term encourages the agent to revise prompts and progressively improve synthesis quality.

\textbf{Behavior Reward.}
We unify tool correctness, format validity, and efficiency into a single objective:
\begin{equation}
\begin{aligned}
R_{\text{beh}} =
& \sum_{t=1}^{T} ( \Phi(a_{t-1}, a_t)
+ \lambda_{KR} \mathbb{I}[a_t = KR \land S_t < \delta]) \\
& + \mathbb{I}[\text{format}(\hat{y})]
- \lambda_T \max(0, T - T_{\max}),
\end{aligned}
\end{equation}
where $a_t$ is the action at turn $t$, and $T$ is the total number of turns. The function $\Phi(a_{t-1}, a_t)$ is a penalty term that enforces valid tool transition rules (e.g., \textit{PG} $\rightarrow$ \textit{IG} $\rightarrow$ \textit{QE}), preventing illogical tool sequences. The knowledge-driven term, weighted by $\lambda_{\text{KR}}$, encourages the agent to invoke KR when the current synthesis quality $S_t$ falls below a threshold $\delta$. Furthermore, $\mathbb{I}[\text{format}(\hat{y})]$ enforces structural validity of the generated output, while the final term penalizes overly long trajectories beyond a predefined limit $T_{\max}$.

\noindent\textbf{Discussion.}
Together, the three rewards form a closed-loop signal: $R_{\text{task}}$ evaluates outcomes, $R_{\text{ref}}$ enables iterative self-improvement, and $R_{\text{beh}}$ regularizes decisions. This design transforms anomaly synthesis from a single-step generation task into a process driven by self-reflection, knowledge retrieval, and iterative refinement, enabling the agent to progressively refine its outputs while maintaining coherent and efficient tool-use strategies.

\section{Experiments}

\subsection{Experiment Settings}
\textbf{Datasets and Trajectories}.
We conduct experiments on two industrial anomaly detection benchmarks, MVTec-AD~\cite{DBLP:journals/ijcv/BergmannBFSS21} and VisA~\cite{DBLP:conf/eccv/ZouJPZD22}. MVTec-AD contains 15 categories with high-resolution images and 1--7 anomaly types per category. Following prior work~\cite{DBLP:conf/aaai/HuZYDCLWW24, DBLP:conf/cvpr/Zhao25}, we use 1/3 of images for training and the remaining 2/3 for testing. VisA includes 12 categories with complex scenes and 1200 real-world defect images. Based on anomaly images from the VisA dataset, we construct multi-turn trajectories using \emph{Gemini 3.1 Pro} and \emph{Gemini 3.1 Flash Image Preview}, resulting in 2772 structured trajectories for SFT (2400/360 with/without KR/12) and 1030 initial prompts for GRPO training. \textbf{Note that our training data has no  overlap with the anomaly types in the test set.}

\noindent\textbf{Evaluation Metrics.}
We evaluate AnomalyAgent in two ways: anomaly generation quality and downstream task performance. For anomaly generation, we use the Inception Score \textbf{(IS)} to evaluate generation quality, and Intra-cluster pairwise LPIPS distance \textbf{(IC-L)} to measure the generation diversity. For downstream tasks, we train a ResNet34 anomaly detection model on data directly generated by AnomalyAgent to evaluate classification accuracy, and train a simple UNet model to evaluate pixel-level and image-level metrics, including \textbf{AUROC}, \textbf{AP}, and \textbf{F1-score}.

\noindent\textbf{Compared Methods.}
We compare AnomalyAgent with representative zero-shot anomaly synthesis methods. For zero-shot approaches, we consider CutPaste~\cite{DBLP:conf/cvpr/LiSYP21}, DRAEM~\cite{DBLP:conf/iccv/ZavrtanikKS21}, NSA~\cite{DBLP:conf/eccv/SchluterTHK22}, and RealNet~\cite{DBLP:conf/cvpr/Zhang0Z24}, as well as recent generative methods including AnomalyAny~\cite{DBLP:conf/cvpr/SunCDF25}, AnoStyler~\cite{DBLP:journals/corr/abs-2511-06687}, and AnoHybrid~\cite{DBLP:conf/cvpr/Zhao25}. We further compare with three powerful image generation and editing models, including \emph{Gemini 3.1 Flash Image Preview}, \emph{GPT Image 1.5}, and \emph{Grok Imagine Image}, which generate anomaly images using fixed prompts, with additional details provided in the supplementary materials. All methods are evaluated under the same protocol on MVTec-AD.

\noindent\textbf{Implementation Details.}
AnomalyAgent is built upon the Qwen3-VL-4B-Thinking backbone\cite{DBLP:journals/corr/abs-2511-21631}. 
IG uses \emph{Gemini 3.1 Flash Image Preview (Nano Banana 2)}, PG and QE use \emph{Gemini 3.1 Pro}, KR leverages \emph{Google Search}, and MG is implemented with a pre-trained MetaUAS\cite{DBLP:conf/nips/Gao24} model. 
In SFT, we freeze the vision encoder and train the multimodal projector and language model using AdamW with cosine decay in bfloat16 under DeepSpeed ZeRO-3 for 3 epochs (lr=$1\times10^{-5}$). In RL, we adopt GRPO with 8 rollouts per prompt and a replay buffer of 128, applying temperature 1.0 and zero-advantage filtering, and train for 2 epochs with ZeRO-3 and offloaded Adam.

\begin{table}[t]
\centering
\caption{Comparison of IS and IC-L on the MVTec dataset. Our method achieves the best IS and IC-L scores. Bold and underline represent optimal and sub-optimal results.}
\label{tab:isil}
\small
\setlength{\tabcolsep}{8pt} % 2. 压缩列间距
\begin{tabular}{l|c|c}
\toprule
\textbf{Method} & \textbf{IS$\uparrow$} & \textbf{IC-L$\uparrow$} \\
\midrule
\multicolumn{3}{c}{\textbf{\textit{Traditional Zero-Shot Methods}}} \\
\cmidrule(lr){1-3}
CutPaste\cite{DBLP:conf/cvpr/LiSYP21} & 1.76 & 0.22 \\
DRAEM\cite{DBLP:conf/iccv/ZavrtanikKS21} & 1.76 & 0.25 \\
NSA\cite{DBLP:conf/eccv/SchluterTHK22} & 1.44 & 0.26 \\
RealNet\cite{DBLP:conf/cvpr/Zhang0Z24} & 1.64 & 0.22 \\
AnomalyAny\cite{DBLP:conf/cvpr/SunCDF25} & 2.02 & \textbf{0.33} \\
AnoStyler\cite{DBLP:journals/corr/abs-2511-06687} & 2.04 & \underline{0.32} \\
AnoHybrid\cite{DBLP:conf/cvpr/Zhao25} & \underline{2.06} & \underline{0.32} \\
\midrule
\multicolumn{3}{c}{\textbf{\textit{Image Generation and Editing Models}}} \\
\cmidrule(lr){1-3}
Gemini 3.1 Flash Image Preview & 1.91 & 0.29 \\
GPT Image 1.5 & 1.77 & 0.29 \\
Grok Imagine Image & 1.68 & 0.28 \\
\midrule
\rowcolor{cyan!5}
\textbf{AnomalyAgent (Ours)} & \textbf{2.10} & \textbf{0.33} \\
\bottomrule
\end{tabular}
\end{table}

\begin{table}[t]
\centering
\caption{Comparison of average anomaly classification accuracy on MVTec-AD. Our method achieves the highest mean performance.}
\label{tab:classification}
\small
\setlength{\tabcolsep}{2pt} % 增加间距使表格更舒展
\begin{tabular}{l|c}
\toprule
\textbf{Method} & \textbf{Accuracy $\uparrow$} \\
\midrule
\multicolumn{2}{c}{\textbf{\textit{Traditional Zero-Shot Methods}}} \\
\cmidrule(lr){1-2}
AnoStyler\cite{DBLP:journals/corr/abs-2511-06687} & 32.2 \\
AnoHybrid\cite{DBLP:conf/cvpr/Zhao25}      & \underline{52.6} \\
\midrule
\multicolumn{2}{c}{\textbf{\textit{Image Generation and Editing Models}}} \\
\cmidrule(lr){1-2}
Gemini 3.1 Flash Image Preview & 44.7 \\
GPT Image 1.5 & 40.5 \\
Grok Imagine Image & 38.9 \\
\midrule
\rowcolor{cyan!5} % 如果你用了 xcolor 宏包，可以给自家的模型加个浅色底色
\textbf{AnomalyAgent (Ours)}                & \textbf{57.0} \\
\bottomrule
\end{tabular}
\end{table}
\begin{table*}[t]
\centering
\caption{Comprehensive performance comparison on MVTec-AD dataset. The table is structured to compare Pixel-Level (left) and Image-Level (right) performance across four methods. \textbf{Bold} indicates the best results. (Gemini-img: Gemini 3.1 Flash Image Preview; GPT-img: GPT Image 1.5)}
\label{tab:localization}
\setlength{\tabcolsep}{1.5pt} 
\resizebox{\textwidth}{!}{ 
\begin{tabular}{l | ccc | ccc | ccc | ccc || ccc | ccc | ccc | ccc}
\toprule
% 第一层：方法名
\multirow{3}{*}{\textbf{Category}} & \multicolumn{3}{c|}{AnoHybrid\cite{DBLP:conf/cvpr/Zhao25}} & \multicolumn{3}{c|}{Gemini-img} & \multicolumn{3}{c|}{GPT-img} & \multicolumn{3}{c||}{\textbf{AnomalyAgent}} & \multicolumn{3}{c|}{AnoHybrid\cite{DBLP:conf/cvpr/Zhao25}} & \multicolumn{3}{c|}{Gemini-img} & \multicolumn{3}{c|}{GPT-img} & \multicolumn{3}{c}{\textbf{AnomalyAgent}} \\
\cmidrule(lr){2-4} \cmidrule(lr){5-7} \cmidrule(lr){8-10} \cmidrule(lr){11-13} \cmidrule(lr){14-16} \cmidrule(lr){17-19} \cmidrule(lr){20-22} \cmidrule(lr){23-25}

% 第二层：评估维度 (Pixel vs Image)
 & \multicolumn{12}{c||}{\textbf{Pixel-Level Performance}} & \multicolumn{12}{c}{\textbf{Image-Level Performance}} \\
\cmidrule(lr){2-13} \cmidrule(lr){14-25}

% 第三层：指标名
 & AUC & AP & $F_1$ & AUC & AP & $F_1$ & AUC & AP & $F_1$ & AUC & AP & $F_1$ & AUC & AP & $F_1$ & AUC & AP & $F_1$ & AUC & AP & $F_1$ & AUC & AP & $F_1$ \\
\midrule
bottle     & 98.3 & \textbf{77.2} & \textbf{72.5} & 96.1 & 74.5 & 68.1 & 86.0 & 52.7 & 51.5 & \textbf{97.2} & 73.5 & 69.7 & 99.2 & \textbf{99.8} & \textbf{98.7} & 98.7 & 99.5 & 97.7 & \textbf{99.5} & \textbf{99.8} & 97.7 & 98.7 & 99.6 & 98.1 \\
cable      & 94.1 & 76.0 & \textbf{73.4} & 93.3 & 68.5 & 60.2 & 90.2 & 55.5 & 50.4 & \textbf{97.4} & \textbf{76.5} & 69.2 & 96.1 & 97.7 & 91.9 & 98.8 & 98.9 & 94.7 & 98.3 & 98.8 & \textbf{97.4} & \textbf{99.1} & \textbf{99.3} & 96.1 \\
capsule    & \textbf{98.4} & \textbf{51.7} & \textbf{54.6} & 96.0 & 44.0 & 47.3 & 88.3 & 33.2 & 34.5 & 96.6 & 46.0 & 49.7 & \textbf{94.9} & \textbf{98.8} & \textbf{95.2} & 90.8 & 96.7 & 92.0 & 90.2 & 91.3 & 88.7 & 91.7 & 97.6 & 91.8 \\
carpet     & 98.6 & 82.8 & 75.6 & 98.8 & 79.0 & 72.3 & 91.3 & 69.0 & 60.4 & \textbf{99.4} & \textbf{85.1} & \textbf{76.4} & 96.3 & 98.9 & 94.0 & 97.9 & 99.2 & 96.0 & 95.6 & 96.1 & 95.9 & \textbf{99.4} & \textbf{99.8} & \textbf{98.4} \\
grid       & \textbf{98.8} & \textbf{58.6} & \textbf{59.2} & 97.4 & 27.5 & 38.6 & 89.0 & 22.1 & 32.7 & 97.8 & 40.3 & 44.9 & \textbf{100} & \textbf{100} & \textbf{100} & 95.6 & 98.0 & 93.7 & 95.7 & 98.0 & 93.7 & \textbf{100} & \textbf{100} & \textbf{100} \\
hazelnut   & \textbf{99.6} & \textbf{89.4} & \textbf{82.6} & 96.7 & 81.1 & 77.4 & 90.3 & 78.3 & 70.2 & 99.4 & 86.4 & 80.1 & 96.7 & 98.3 & 92.6 & 96.2 & 97.0 & 90.4 & 94.1 & 97.9 & \textbf{96.0} & \textbf{98.9} & \textbf{99.2} & 95.9 \\
leather    & 99.6 & 72.7 & 67.1 & 99.4 & 72.8 & 68.3 & 93.9 & 69.4 & 65.6 & \textbf{99.7} & \textbf{80.3} & \textbf{73.9} & 98.4 & 99.5 & 97.4 & \textbf{100} & \textbf{100} & \textbf{100} & 96.9 & 98.0 & 95.1 & \textbf{100} & \textbf{100} & \textbf{100} \\
metal nut  & 98.8 & 93.5 & 87.0 & 97.5 & 90.6 & 83.6 & 86.2 & 58.5 & 58.6 & \textbf{99.3} & \textbf{95.4} & \textbf{88.5} & \textbf{99.8} & \textbf{99.9} & \textbf{99.2} & 99.4 & 99.8 & 97.7 & 95.5 & 98.6 & 93.5 & \textbf{99.8} & \textbf{99.9} & 98.7 \\
pill       & 99.3 & \textbf{94.9} & \textbf{88.4} & 99.1 & 82.7 & 72.7 & 88.0 & 65.1 & 60.3 & \textbf{99.5} & 84.0 & 80.4 & \textbf{99.1} & \textbf{99.8} & \textbf{98.9} & 92.3 & 98.0 & 93.0 & 92.5 & 96.0 & 91.9 & 97.4 & 98.8 & 96.6 \\
screw      & 77.0 & 7.8 & 6.4 & 97.6 & 10.5 & 17.3 & 94.7 & 21.6 & 29.9 & \textbf{98.4} & \textbf{55.4} & \textbf{57.8} & 44.6 & 72.6 & 84.9 & 88.4 & 94.5 & 88.1 & 90.9 & 95.3 & 88.6 & \textbf{94.6} & \textbf{97.8} & \textbf{91.3} \\
tile       & 99.3 & \textbf{94.6} & \textbf{87.4} & 97.3 & 92.6 & 80.5 & 94.3 & 59.0 & 59.9 & \textbf{99.3} & \textbf{94.6} & 85.7 & 99.5 & 99.8 & 99.0 & 99.3 & 99.6 & 97.4 & 98.8 & 99.5 & 95.0 & \textbf{100} & \textbf{100} & \textbf{100} \\
toothbrush & 98.7 & 65.2 & 67.8 & \textbf{99.5} & \textbf{73.7} & \textbf{75.4} & 94.8 & 31.1 & 42.6 & 99.3 & 72.3 & 70.3 & \textbf{100} & \textbf{100} & \textbf{100} & \textbf{100} & \textbf{100} & \textbf{100} & 98.3 & 99.1 & 95.2 & \textbf{100} & \textbf{100} & \textbf{100} \\
transistor & \textbf{98.1} & \textbf{80.8} & \textbf{74.2} & 90.6 & 62.1 & 59.9 & 88.2 & 60.3 & 54.7 & 90.5 & 63.7 & 62.1 & 92.9 & 90.4 & 86.3 & 94.5 & 92.6 & 90.9 & 95.9 & \textbf{97.9} & \textbf{93.4} & \textbf{98.3} & 97.1 & 92.9 \\
wood       & 95.8 & 70.7 & 64.8 & 95.8 & 78.5 & 72.8 & 82.2 & 45.6 & 49.0 & \textbf{97.5} & \textbf{79.8} & \textbf{74.0} & 96.6 & 98.7 & 98.7 & 98.5 & 99.3 & 97.7 & 97.3 & 98.0 & 94.3 & \textbf{100} & \textbf{100} & \textbf{100} \\
zipper     & \textbf{99.1} & \textbf{82.3} & \textbf{74.9} & 96.9 & 75.8 & 66.4 & 90.4 & 55.0 & 49.4 & 98.9 & 79.8 & 72.4 & \textbf{99.9} & \textbf{100} & 99.3 & 99.2 & 99.7 & 97.0 & 96.3 & 98.1 & 94.2 & \textbf{99.9} & \textbf{100} & \textbf{99.4} \\
\midrule
\textbf{Mean} & 96.9 & 72.9 & 69.1 & 96.8 & 67.6 & 64.1 & 89.9 & 51.8 & 51.3 & \textbf{98.0} & \textbf{74.2} & \textbf{70.3} & 94.3 & 96.9 & 95.7 & 96.6 & 98.2 & 95.1 & 95.7 & 97.5 & 94.0 & \textbf{98.5} & \textbf{99.3} & \textbf{97.3} \\
\bottomrule
\end{tabular}
}
\end{table*}

\subsection{Main Results}
We evaluate AnomalyAgent on the MVTec-AD datasets across three dimensions: image generation quality, downstream anomaly classification, and downstream anomaly localization.

\noindent\textbf{Anomaly Generation Results.}
As shown in Table~\ref{tab:isil}, AnomalyAgent achieves a mean IS of 2.10 and IC-L of 0.33 on the MvTec-AD dataset, surpassing previous zero-shot SOTA methods. The higher IS reflects improved structural fidelity and anomaly recognizability under agent-based planning. Meanwhile, the competitive IC-L indicates that the ``Planner-Executor-Validator'' loop effectively explores the anomaly space, producing diverse defect patterns and mitigating mode collapse.

\noindent\textbf{Anomaly Classification Results.}
As shown in Table~\ref{tab:classification}, our method achieves the best overall performance with a mean accuracy of 57.0\%, outperforming all competing approaches by a clear margin. In particular, AnomalyAgent surpasses the strongest traditional zero-shot baseline, AnoHybrid, by +4.4\%, and significantly outperforms image generation and editing models. Compared with recent traditional zero-shot methods (e.g., AnoStyler and AnoHybrid), which typically rely on single-step generation, our approach leverages iterative reasoning and multi-turn tool interaction to produce more semantically consistent and diverse anomalies, leading to more informative supervision signals. Furthermore, while image generation and editing models demonstrate reasonable performance, their lack of task-specific optimization limits their effectiveness. In contrast, our agent-based framework enables closed-loop refinement and better alignment with anomaly characteristics, resulting in superior classification performance.

\noindent\textbf{Anomaly Detection Results.}
Table~\ref{tab:localization} presents a comprehensive comparison on the MVTec-AD dataset for anomaly localization. Our AnomalyAgent achieves the best overall performance across all metrics, demonstrating consistent advantages at both pixel-level and image-level. Compared with the strongest traditional zero-shot baseline, AnoHybrid\cite{DBLP:conf/cvpr/Zhao25}, our method yields consistent improvements, particularly in pixel-level metrics. Specifically, AnomalyAgent improves the mean pixel-level AUC from 96.9 to 98.0, AP from 72.9 to 74.2, and F1-score from 69.1 to 70.3. These gains indicate that our approach produces more accurate and better-localized anomaly regions. In comparison with image generation and editing models, including \emph{Gemini 3.1 Flash Image Preview} and \emph{GPT Image 1.5}, our method shows a substantial margin across all metrics. While these models can generate plausible anomalies using fixed prompts, their lack of task-specific prompt optimization leads to inferior localization performance. In contrast, our agent-based framework enables iterative refinement and better alignment with anomaly structures, resulting in significantly improved detection accuracy.

\begin{figure}
    \centering
    \includegraphics[width=\linewidth]{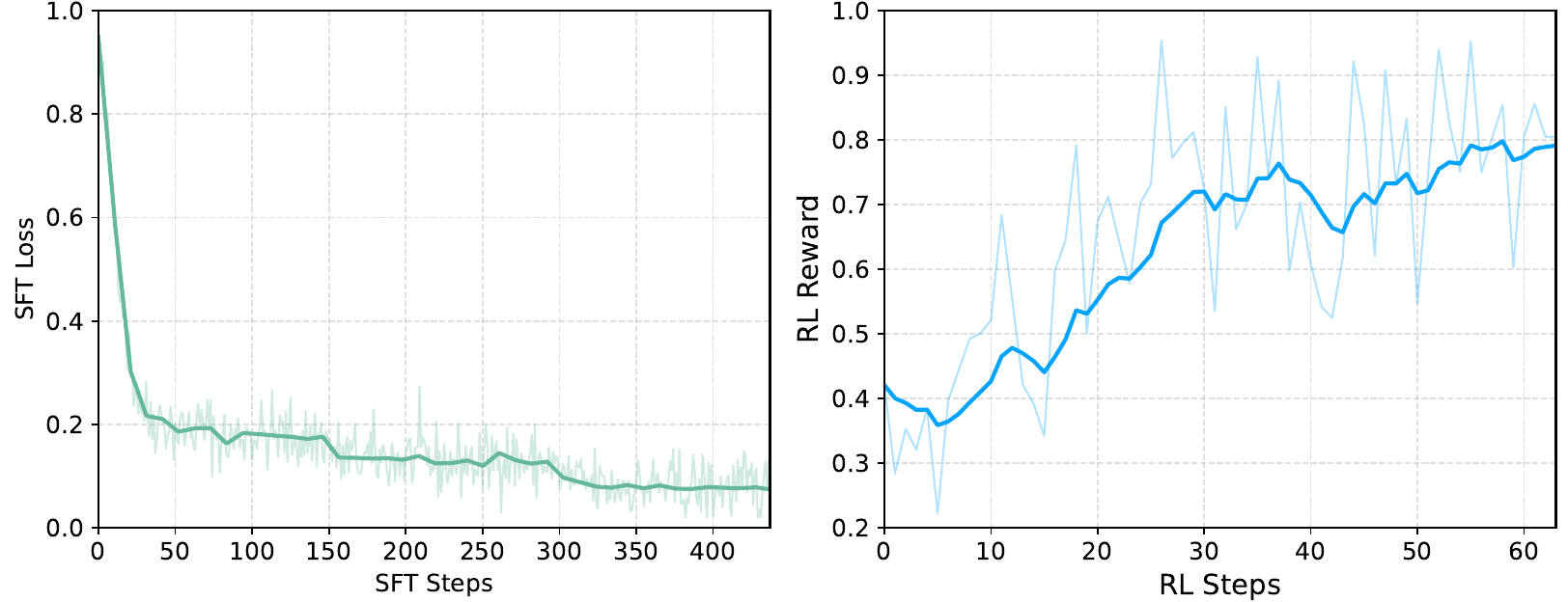}
    \caption{Training dynamics of AnomalyAgent. \textbf{Left:} SFT loss rapidly decreases and converges. \textbf{Right:} RL reward steadily improves throughout GRPO optimization.}
    
    \label{fig:curve}
\end{figure}

\begin{figure*}[t]
  \centering
  \includegraphics[width=0.71\linewidth]{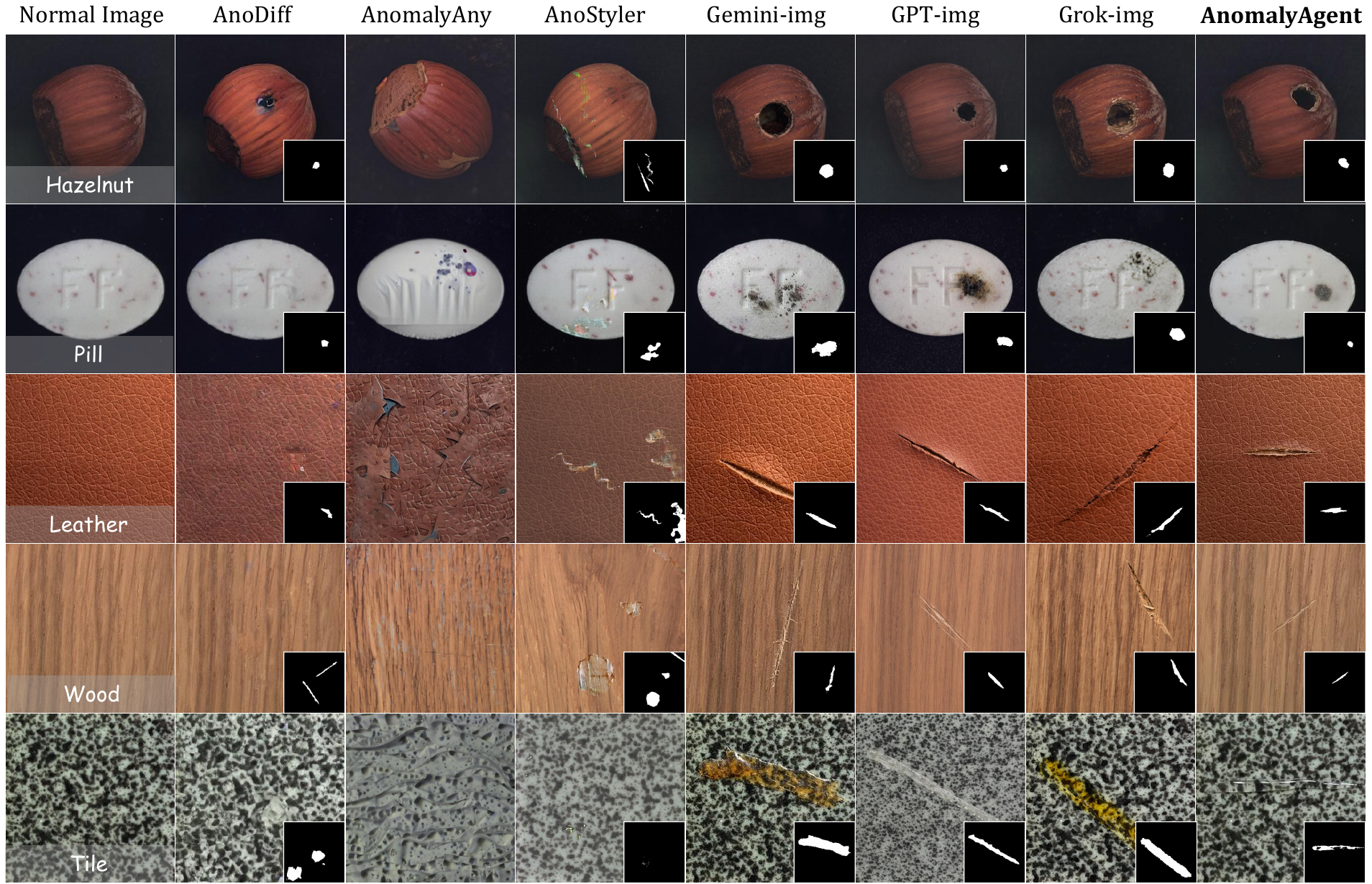}
  \caption{Visualization of anomaly synthesis results on MVTec-AD. AnomalyAgent achieves higher semantic consistency and more accurate defect localization than traditional generative models (Gemini-img: Gemini 3.1 Flash Image Preview; GPT-img: GPT Image 1.5; Grok-img: Grok Imagine Image). }
  \Description{visual Comparison.}
  \label{fig:vis_compare}
\end{figure*}

\subsection{Training Dynamics}
As shown in Fig.\ref{fig:curve}, the left vertical axis shows the SFT loss, and the right vertical axis shows the RL reward over training steps. During the SFT stage (first 300 steps), the loss decreases rapidly from 0.94 to 0.10, indicating that the model effectively learns to imitate the structured trajectories. After switching to RL with GRPO, the reward steadily increases from 0.42 to 0.79, reflecting progressive improvement in generation quality, tool-use efficiency, and self-reflection capability. Notably, the reward saturates after approximately 50 global steps, suggesting convergence of the agentic policy. This complementary trend validates our two-stage paradigm, where cold-start SFT provides a reliable behavioral foundation and RL further unlocks the potential of agent through targeted optimization.

\subsection{Visualization Results}
Fig. \ref{fig:vis_compare} compares AnomalyAgent with representative synthesis methods. Compared to AnoDiff\cite{DBLP:conf/aaai/HuZYDCLWW24}, which relies on constrained real-world samples, AnomalyAgent generates images with competitive visual realism. While zero-shot models (e.g., AnoStyler\cite{DBLP:journals/corr/abs-2511-06687}, AnomalyAny\cite{DBLP:conf/cvpr/SunCDF25}) generalize well, they often lack fine-grained realism. Furthermore, generating images directly from image generation and editing models often results in semantic inconsistencies with the underlying texture, requiring subtle prompts to unlock their potential. In contrast, \textbf{AnomalyAgent} synthesizes defects with high visual fidelity and accurate spatial localization. By framing synthesis as a multi-turn decision process, our agent leverages iterative tool-driven refinement to effectively mitigate artifacts like semantic drift and incoherent boundaries. These results demonstrate the clear advantage of agentic reasoning over traditional single-step generation in complex industrial scenarios.

\subsection{Ablation Studies}
We conduct ablation studies to analyze the contribution of each component in AnomalyAgent.

\begin{table}[t]
\centering
\caption{Ablation study of proposed components with different task categories.}
\label{tab:ablation_1}

\footnotesize % 或者使用 \scriptsize 让表格更小
\setlength{\tabcolsep}{4pt} % 进一步缩小列间距（原先是4pt）
    \begin{tabular}{c|ccccc|cc|c}
    \toprule
    \multirow{2}{*}{\#} & \multicolumn{5}{c|}{Method} & \multicolumn{2}{c|}{Generation} & Classification \\
    \cmidrule(lr){2-6} \cmidrule(lr){7-8} \cmidrule(lr){9-9}
    & PG & QE & KR & SFT & RL & IS $\uparrow$ & IC-L $\uparrow$ & Accuracy $\uparrow$ \\
    \midrule
    (a) &            &            &            &            &            & 1.91 & 0.29 & 44.7 \\
    (b) &  &  \checkmark  &  \checkmark  &            &            & 1.90 & 0.29 & 48.2 \\
    (c) & \checkmark & \checkmark &  &            &            & 2.03 & 0.32 & 47.3 \\
    (d) & \checkmark &  &  &            &            & 1.99 & 0.32 & 45.1 \\
    (e) & \checkmark & \checkmark & \checkmark &  &            & 2.03 & 0.32 & 49.5 \\
    (f) & \checkmark & \checkmark & \checkmark & \checkmark &            & 2.08 & 0.33 & 52.4 \\
    \rowcolor{cyan!5} 
    (g) & \checkmark & \checkmark & \checkmark & \checkmark & \checkmark & 2.10 & 0.33 & 57.0 \\
    \bottomrule
    \end{tabular}

\end{table}
\noindent\textbf{Component Analysis.}
As shown in Table~\ref{tab:ablation_1}, starting from a minimal setup (a), progressively enabling key modules consistently improves performance. PG and QE significantly improve accuracy from 44.7\% to 47.3\%, demonstrating the importance of aligning prompts with visual defects and incorporating feedback. KR further improves classification accuracy (49.5\%), indicating that external semantic context helps refine anomaly realism. Adding SFT stabilizes tool-use behavior and yields consistent gains (52.4\%), while RL provides the largest improvement, boosting accuracy to 57.0\% by enabling iterative optimization and long-horizon decision-making. Overall, all components contribute cumulatively, validating the effectiveness of the proposed agentic pipeline.

\begin{table}[t]
\centering
\caption{Ablation study of different reward components.}
\label{tab:ablation_reward}
\small
    \setlength{\tabcolsep}{4pt}
    \begin{tabular}{c|ccc|cc|c}
    \toprule
    \multirow{2}{*}{\#} & \multicolumn{3}{c|}{Rewards} & \multicolumn{2}{c|}{Generation} & Classification \\
    \cmidrule(lr){2-4} \cmidrule(lr){5-6} \cmidrule(lr){7-7}
    & $R_{\text{task}}$ & $R_{\text{ref}}$ & $R_{\text{beh}}$ & IS $\uparrow$ & IC-L $\uparrow$ & Accuracy $\uparrow$ \\
    \midrule
    (a) &  &  &  & 2.08 & 0.33 & 52.4 \\
    (b) & \checkmark &  &  & 2.09 & 0.33 & 53.6 \\
    (c) & \checkmark & \checkmark &  & 2.09 & 0.33 & 55.3 \\
    \rowcolor{cyan!5} 
    (d) & \checkmark & \checkmark & \checkmark & 2.10 & 0.33 & 57.0 \\
    \bottomrule
    \end{tabular}

\end{table}

\noindent\textbf{Reward Analysis.}
Table~\ref{tab:ablation_reward} further examines the impact of different reward components. Using only the task reward already improves performance over the SFT baseline (52.4\% $\rightarrow$ 53.6\%), indicating that optimizing final output quality is effective. Incorporating the reflection reward brings additional gains (55.3\%), demonstrating its role in encouraging iterative refinement. Finally, adding the behavior reward yields the best performance (57.0\%), confirming that regularizing tool-use correctness and efficiency is critical for stable multi-turn reasoning. These results highlight that the three rewards are complementary, jointly promoting high-quality generation, progressive improvement, and disciplined decision-making.

\section{Conclusion}
In this paper, we propose \emph{AnomalyAgent}, a novel agent-based framework that formulates industrial anomaly synthesis as a multi-turn decision-making process. By integrating structured tool use with iterative reasoning, the proposed method enables controllable and semantically consistent anomaly generation beyond conventional single-step pipelines. We further design a two-stage training strategy: SFT based on the constructed synthetic trajectory, followed by optimization of long-range decision-making capabilities through RL. Extensive experiments demonstrate that AnomalyAgent outperforms existing zero-shot methods in both anomaly generation quality and downstream tasks. Future work will explore scaling to more complex industrial scenarios and extending the agent framework to broader multimodal generation tasks.


% \begin{acks}
% This work was supported by industrial partners and university grants. We thank all contributors for helpful discussions and technical support.
% \end{acks}

\bibliographystyle{ACM-Reference-Format}
\bibliography{AnomalyAgent}

\clearpage

\appendix
% \onecolumn
% \section*{Appendices} % 附录的总标题
% \addcontentsline{toc}{section}{Appendices} % 将“附录”加入主目录

% % 建立局部目录结构
% \startcontents[appendices]
% \printcontents[appendices]{l}{1}{\setcounter{tocdepth}{2}} 
% \vspace{1cm}
% \hrule
% \vspace{1cm}

\noindent\begin{minipage}{\textwidth}

\section{Cost Analysis}

We conduct a comprehensive efficiency analysis of AnomalyAgent, focusing on generation quality, inference time, and monetary cost.

\noindent\textbf{Human Evaluation Protocol.} To ensure a rigorous assessment of the synthesized anomalies, we three invite industrial anomaly detection experts to perform a blind review. For each method, 150 synthesized images are randomly sampled. Experts are required to score each image on a scale of 1 to 10 based on two key dimensions: Defect Realism, measuring the fidelity of synthesized textures, and Semantic Consistency, reflecting alignment with industrial logic. To filter out samples with partial flaws, an image is classified as a ``Good Sample'' only if it achieves a score of no less than 8 in both dimensions from all experts. The Good Sample Rate is then calculated to represent the high-quality synthesis capability of each method. Monetary cost is calculated based on official Batch API pricing as of April 2026.

\noindent\textbf{Quantitative Efficiency Analysis.} As presented in Table~\ref{tab:efficiency_comparison}, AnomalyAgent achieves the highest success rate of 91.3\% and the lowest time per good sample of 117.2s. Compared to Gemini 3.1 Flash Image Preview, which attains a 77.3\% success rate and 137.8s per good sample, AnomalyAgent improves the success rate by 14.0 percentage points while reducing the time per good sample by 15.0\%. Although AnoStyler achieves the lowest per-request latency at 21.0s, its low success rate of 16.7\% leads to a substantially higher effective cost of 125.7s per good sample, making it less viable for large-scale industrial deployment. In terms of monetary efficiency, AnomalyAgent achieves the lowest cost per good sample at \$0.0567, outperforming Gemini 3.1 Flash Image Preview at \$0.0595 and GPT Image 1.5 at \$0.1354. While Grok Imagine Image offers the lowest per-request price, its lower quality results in a higher effective cost of \$0.0751 compared to our method.

\noindent\textbf{Trade-off Analysis.} Fig.~\ref{fig:cost} further illustrates the trade-offs between efficiency and quality. In both time–quality and cost–quality spaces, our method consistently occupies the most favorable region, achieving the highest good sample rate (91.3\%) while maintaining the lowest effective time and cost. This indicates that our approach dominates all baselines in terms of practical efficiency.

\vspace{2em} 
% 表格 - 使用 \captionof{table}{...} 而不是 table 环境
\centering
\captionof{table}{Comparison of efficiency and cost effectiveness across different methods. The best results are highlighted in \textbf{bold}.}
\label{tab:efficiency_comparison}
\begin{adjustbox}{width=0.7\linewidth}
\begin{tabular}{lccccc}
\toprule
\multirow{2}{*}{\textbf{Method}} & \textbf{Good Sample Rate} & \multicolumn{2}{c}{\textbf{Per Sample}} & \multicolumn{2}{c}{\textbf{Per Good Sample}} \\
\cmidrule(lr){3-4} \cmidrule(lr){5-6}
& (\%) $\uparrow$ & Time (s) $\downarrow$ & Cost (\$) $\downarrow$ & Time (s) $\downarrow$ & Cost (\$) $\downarrow$ \\
\midrule
AnoStyler & 16.7 & \textbf{21.0} & -- & 125.7 & -- \\
AnomalyAny & 34.0 & 250.0 & -- & 735.3 & -- \\
Gemini 3.1 Flash Image Preview& 77.3 & 101.0 & 0.0436 & 137.8 & 0.0595 \\
GPT Image 1.5 & 58.7 & 122.0 & 0.0795 & 207.8 & 0.1354 \\
Grok Imagine Image & 42.7 & 105.0 & \textbf{0.0321} & 245.9 & 0.0751 \\
\midrule
\rowcolor{cyan!5} \textbf{AnomalyAgent (Ours)} & \textbf{91.3} & 107.0 & 0.0518 & \textbf{117.2} & \textbf{0.0567} \\
\bottomrule
\end{tabular}
\end{adjustbox}

\vspace{2em}  % 表格和图片之间的间距

% 图片 - 使用 \captionof{figure}{...}
\centering
\includegraphics[width=\linewidth]{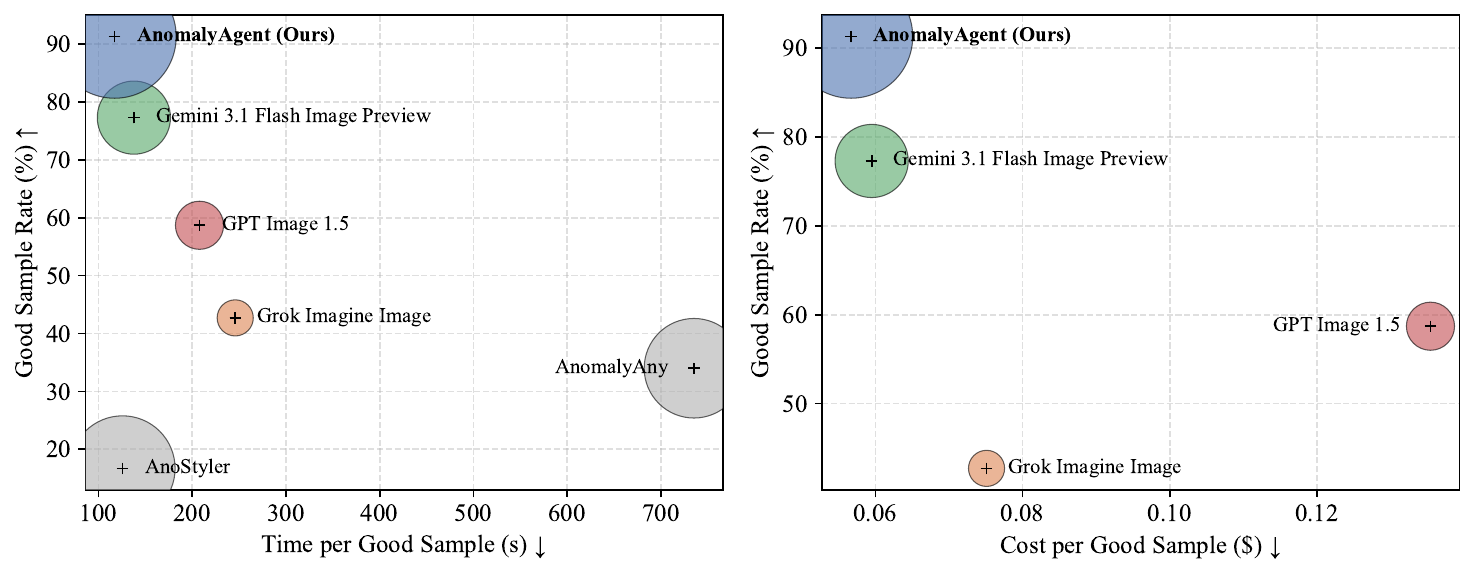}
\captionof{figure}{Comparison of efficiency and cost-effectiveness. Bubble size indicates the Inception Score (IS). Our AnomalyAgent achieves the best balance between quality, time, and cost.}
\label{fig:cost}

\end{minipage}

\clearpage

\noindent\begin{minipage}{\textwidth}

\section{Extended Quantitative Results}

We present an extended category-wise evaluation on MVTec-AD in Tables~\ref{tab:mvtec_image_full} and \ref{tab:mvtec_pixel_full}, comparing AnomalyAgent against a broader range of classical and generative baselines. Our method achieves state-of-the-art performance at both the image level and pixel level. These results demonstrate that the agentic iterative refinement and industrial knowledge retrieval of AnomalyAgent ensure superior defect realism and structural alignment, leading to more precise anomaly localization..

\vspace{2em}  % 表格和图片之间的间距
\centering
\captionof{table}{Comparison of image-level performance (AUC, AP, and $F_1$-max) on MVTec-AD dataset. \textbf{Bold} indicates the best results. (Gemini-img: Gemini 3.1 Flash Image Preview; GPT-img: GPT Image 1.5; Grok-img: Grok Imagine Image)}
\label{tab:mvtec_image_full}
\setlength{\tabcolsep}{2.2pt} 
\resizebox{\textwidth}{!}{ 
\begin{tabular}{l | ccc | ccc | ccc | ccc | ccc | ccc | ccc | ccc}
\toprule
\textbf{Category} & \multicolumn{3}{c|}{CropPaste} & \multicolumn{3}{c|}{DRAEM} & \multicolumn{3}{c|}{DFMGAN} & \multicolumn{3}{c|}{AnoHybrid} & \multicolumn{3}{c|}{Gemini-img} & \multicolumn{3}{c|}{GPT-img} & \multicolumn{3}{c|}{Grok-img} & \multicolumn{3}{c}{\textbf{AnomalyAgent}} \\
\cmidrule(lr){2-4} \cmidrule(lr){5-7} \cmidrule(lr){8-10} \cmidrule(lr){11-13} \cmidrule(lr){14-16} \cmidrule(lr){17-19} \cmidrule(lr){20-22} \cmidrule(lr){23-25}

 & AUC & AP & $F_1$ & AUC & AP & $F_1$ & AUC & AP & $F_1$ & AUC & AP & $F_1$ & AUC & AP & $F_1$ & AUC & AP & $F_1$ & AUC & AP & $F_1$ & AUC & AP & $F_1$ \\
\midrule
bottle     & 85.4 & 95.1 & 90.9 & \textbf{99.3} & \textbf{99.8} & \textbf{98.9} & \textbf{99.3} & \textbf{99.8} & 97.7 & 99.2 & \textbf{99.8} & 98.7 & 98.7 & 99.5 & 97.7 & \textbf{99.5} & \textbf{99.8} & 97.7 & 98.0 & 99.3 & 98.8 & 98.7 & 99.6 & 98.1 \\
cable      & 93.3 & 96.1 & 91.6 & 72.1 & 83.2 & 79.2 & 95.9 & 97.8 & 93.8 & 96.1 & 97.7 & 91.9 & 98.8 & 98.9 & 94.7 & 98.3 & 98.8 & \textbf{97.4} & 97.2 & 97.4 & 96.2 & \textbf{99.1} & \textbf{99.3} & 96.1 \\
capsule    & 77.1 & 94.1 & 90.4 & 93.2 & 98.7 & 94.0 & 92.8 & 98.5 & 94.5 & 94.9 & 98.8 & 95.2 & 90.8 & 96.7 & 92.0 & 90.2 & 91.3 & 88.7 & 88.9 & 90.3 & 87.7 & 91.7 & \textbf{97.6} & 91.8 \\
carpet     & 57.7 & 84.3 & 87.3 & 95.3 & 98.7 & 93.4 & 67.9 & 87.9 & 87.3 & 96.3 & 98.9 & 94.0 & 97.9 & 99.2 & 96.0 & 95.6 & 96.1 & 95.9 & 93.6 & 95.7 & 93.9 & \textbf{99.4} & \textbf{99.8} & \textbf{98.4} \\
grid       & 83.0 & 94.1 & 87.6 & 99.8 & 99.9 & 98.8 & 73.0 & 90.4 & 85.4 & \textbf{100} & \textbf{100} & \textbf{100} & 95.6 & 98.0 & 93.7 & 95.7 & 98.0 & 93.7 & 97.5 & 98.7 & 95.2 & \textbf{100} & \textbf{100} & \textbf{100} \\
hazelnut   & 68.8 & 85.0 & 78.0 & \textbf{100} & \textbf{100} & \textbf{100} & 99.9 & \textbf{100} & 99.0 & 96.7 & 98.3 & 92.6 & 96.2 & 97.0 & 90.4 & 94.1 & 97.9 & 96.0 & 92.1 & 95.5 & 95.0 & 98.9 & 99.2 & 95.9 \\
leather    & 91.9 & 97.5 & 90.9 & \textbf{100} & \textbf{100} & \textbf{100} & 99.9 & \textbf{100} & 99.2 & 98.4 & 99.5 & 97.4 & \textbf{100} & \textbf{100} & \textbf{100} & 96.9 & 98.0 & 95.1 & 95.6 & 97.3 & 94.1 & \textbf{100} & \textbf{100} & \textbf{100} \\
metal nut  & 92.2 & 98.1 & 93.3 & 97.8 & 99.6 & 97.6 & 99.3 & 99.8 & \textbf{99.2} & 99.8 & 99.9 & \textbf{99.2} & 99.4 & 99.8 & 97.7 & 95.5 & 98.6 & 93.5 & 97.4 & 99.2 & 95.1 & \textbf{99.8} & \textbf{99.9} & 98.7 \\
pill       & 51.7 & 87.1 & 91.4 & 94.4 & 98.9 & 95.8 & 68.7 & 91.7 & 91.4 & \textbf{99.1} & \textbf{99.8} & \textbf{98.9} & 92.3 & 98.0 & 93.0 & 92.5 & 96.0 & 91.9 & 93.5 & 97.0 & 93.4 & 97.4 & 98.8 & 96.6 \\
screw      & 59.3 & 81.9 & 86.0 & 88.5 & 96.3 & 89.3 & 22.3 & 64.7 & 85.3 & 44.6 & 72.6 & 84.9 & 88.4 & 94.5 & 88.1 & 90.9 & 95.3 & 88.6 & 78.1 & 86.7 & 83.1 & \textbf{94.6} & \textbf{97.8} & \textbf{91.3} \\
tile       & 73.8 & 91.1 & 83.8 & \textbf{100} & \textbf{100} & \textbf{100} & \textbf{100} & \textbf{100} & \textbf{100} & 99.5 & 99.8 & 99.0 & 99.3 & 99.6 & 97.4 & 98.8 & 99.5 & 95.0 & 98.9 & 99.4 & 94.6 & \textbf{100} & \textbf{100} & \textbf{100} \\
toothbrush & 81.2 & 91.0 & 88.9 & \textbf{100} & \textbf{100} & \textbf{100} & \textbf{100} & \textbf{100} & \textbf{100} & \textbf{100} & \textbf{100} & \textbf{100} & \textbf{100} & \textbf{100} & \textbf{100} & 98.3 & 99.1 & 95.2 & 82.5 & 90.7 & 84.2 & \textbf{100} & \textbf{100} & \textbf{100} \\
transistor & 85.9 & 81.8 & 80.0 & 79.6 & 80.5 & 71.4 & 90.8 & 92.5 & 88.9 & 92.9 & 90.4 & 86.3 & 94.5 & 92.6 & 90.9 & 95.9 & \textbf{97.9} & 93.4 & 96.2 & \textbf{98.3} & \textbf{94.4} & \textbf{98.3} & 97.1 & 92.9 \\
wood       & 49.5 & 81.2 & 86.6 & \textbf{100} & \textbf{100} & \textbf{100} & 98.4 & 99.4 & 98.8 & 96.6 & 98.7 & 98.7 & 98.5 & 99.3 & 97.7 & 97.3 & 98.0 & 94.3 & 97.6 & 98.5 & 93.1 & \textbf{100} & \textbf{100} & \textbf{100} \\
zipper     & 59.4 & 82.8 & 88.9 & \textbf{100} & \textbf{100} & \textbf{100} & 99.7 & 99.9 & 99.4 & 99.9 & \textbf{100} & 99.3 & 99.2 & 99.7 & 97.0 & 96.3 & 98.1 & 94.2 & 96.8 & 98.0 & 93.8 & 99.9 & \textbf{100} & 99.4 \\
\midrule
\textbf{Mean} & 74.0 & 89.4 & 87.7 & 94.6 & 97.0 & 94.4 & 87.2 & 94.8 & 94.7 & 94.3 & 96.9 & 95.7 & 96.6 & 98.2 & 95.1 & 95.7 & 97.5 & 94.0 & 93.6 & 96.1 & 92.8 & \textbf{98.5} & \textbf{99.3} & \textbf{97.3} \\
\bottomrule
\end{tabular}
}

\vspace{2em}  % 表格和图片之间的间距

\centering
\captionof{table}{Comparison of pixel-level performance (AUC, AP, and $F_1$-max) on MVTec-AD dataset. \textbf{Bold} indicates the best results. (Gemini-img: Gemini 3.1 Flash Image Preview; GPT-img: GPT Image 1.5; Grok-img: Grok Imagine Image)}
\label{tab:mvtec_pixel_full}

\setlength{\tabcolsep}{2.2pt} 
\resizebox{\textwidth}{!}{ 
\begin{tabular}{l | ccc | ccc | ccc | ccc | ccc | ccc | ccc | ccc}
\toprule
\textbf{Category} & \multicolumn{3}{c|}{CropPaste} & \multicolumn{3}{c|}{DRAEM} & \multicolumn{3}{c|}{DFMGAN} & \multicolumn{3}{c|}{AnoHybrid} & \multicolumn{3}{c|}{Gemini-img} & \multicolumn{3}{c|}{GPT-img} & \multicolumn{3}{c|}{Grok-img} & \multicolumn{3}{c}{\textbf{AnomalyAgent}} \\
\cmidrule(lr){2-4} \cmidrule(lr){5-7} \cmidrule(lr){8-10} \cmidrule(lr){11-13} \cmidrule(lr){14-16} \cmidrule(lr){17-19} \cmidrule(lr){20-22} \cmidrule(lr){23-25}

 & AUC & AP & $F_1$ & AUC & AP & $F_1$ & AUC & AP & $F_1$ & AUC & AP & $F_1$ & AUC & AP & $F_1$ & AUC & AP & $F_1$ & AUC & AP & $F_1$ & AUC & AP & $F_1$ \\
\midrule
bottle     & 94.5 & 67.4 & 63.5 & 96.7 & 80.2 & 74.0 & \textbf{98.9} & \textbf{90.2} & \textbf{83.9} & 98.3 & 77.2 & 72.5 & 96.1 & 74.5 & 68.1 & 86.0 & 52.7 & 51.5 & 92.9 & 63.3 & 62.5 & 97.2 & 73.5 & 69.7 \\
cable      & 96.0 & 75.3 & 69.3 & 80.3 & 21.8 & 28.3 & 97.2 & \textbf{81.0} & \textbf{75.4} & 94.1 & 76.0 & 73.4 & 93.3 & 68.5 & 60.2 & 90.2 & 55.5 & 50.4 & 92.2 & 58.0 & 55.2 & \textbf{97.4} & 76.5 & 69.2 \\
capsule    & 95.3 & 49.2 & 51.1 & 76.2 & 25.5 & 32.1 & 79.2 & 26.0 & 35.0 & \textbf{98.4} & \textbf{51.7} & \textbf{54.6} & 96.0 & 44.0 & 47.3 & 88.3 & 33.2 & 34.5 & 80.3 & 24.2 & 24.9 & 96.6 & 46.0 & 49.7 \\
carpet     & 83.7 & 36.6 & 39.7 & 92.6 & 43.0 & 41.9 & 90.6 & 33.4 & 38.1 & 98.6 & 82.8 & 75.6 & 98.8 & 79.0 & 72.3 & 91.3 & 69.0 & 60.4 & 91.3 & 69.0 & 60.4 & \textbf{99.4} & \textbf{85.1} & \textbf{76.4} \\
grid       & 84.7 & 13.1 & 22.4 & \textbf{99.1} & \textbf{59.3} & 58.7 & 75.2 & 14.3 & 20.5 & 98.8 & 58.6 & \textbf{59.2} & 97.4 & 27.5 & 38.6 & 89.0 & 22.1 & 32.7 & 89.9 & 21.6 & 28.9 & 97.8 & 40.3 & 44.9 \\
hazelnut   & 88.5 & 38.0 & 42.8 & 98.8 & 73.6 & 68.5 & \textbf{99.7} & \textbf{95.2} & \textbf{89.5} & 99.6 & 89.4 & 82.6 & 96.7 & 81.1 & 77.4 & 90.3 & 78.3 & 70.2 & 93.3 & 79.3 & 74.2 & 99.4 & 86.4 & 80.1 \\
leather    & 97.5 & 76.0 & 70.8 & 98.5 & 67.6 & 65.0 & 98.5 & 68.7 & 66.7 & 99.6 & 72.7 & 67.1 & 99.4 & 72.8 & 68.3 & 93.9 & 69.4 & 65.6 & 95.1 & 77.4 & 69.0 & \textbf{99.7} & \textbf{80.3} & \textbf{73.9} \\
metal nut  & 96.3 & 84.2 & 74.0 & 96.9 & 84.2 & 74.5 & \textbf{99.3} & \textbf{98.1} & \textbf{94.5} & 98.8 & 93.5 & 87.0 & 97.5 & 90.6 & 83.6 & 86.2 & 58.5 & 58.6 & 98.1 & 86.6 & 82.3 & \textbf{99.3} & 95.4 & 88.5 \\
pill       & 81.5 & 17.8 & 24.3 & 95.8 & 45.3 & 53.0 & 81.2 & 67.8 & 72.6 & 99.3 & \textbf{94.9} & \textbf{88.4} & 99.1 & 82.7 & 72.7 & 88.0 & 65.1 & 60.3 & 76.0 & 60.2 & 55.3 & \textbf{99.5} & 84.0 & 80.4 \\
screw      & 93.4 & 31.2 & 36.0 & 91.0 & 30.1 & 35.7 & 58.8 & 2.2 & 5.3 & 77.0 & 7.8 & 6.4 & 97.6 & 10.5 & 17.3 & 94.7 & 21.6 & 29.9 & 91.1 & 8.8 & 15.3 & \textbf{98.4} & \textbf{55.4} & \textbf{57.8} \\
tile       & 94.0 & 79.3 & 74.5 & 98.5 & 93.2 & 87.8 & \textbf{99.5} & \textbf{97.1} & \textbf{91.6} & 99.3 & 94.6 & 87.4 & 97.3 & 92.6 & 80.5 & 94.3 & 59.0 & 59.9 & 93.5 & 33.2 & 47.3 & 99.3 & 94.6 & 85.7 \\
toothbrush & 89.3 & 30.9 & 34.6 & 93.8 & 29.5 & 28.4 & 96.4 & \textbf{75.9} & 72.6 & 98.7 & 65.2 & 67.8 & \textbf{99.5} & 73.7 & \textbf{75.4} & 94.8 & 31.1 & 42.6 & 92.8 & 40.4 & 44.3 & 99.3 & 72.3 & 70.3 \\
transistor & 85.9 & 52.5 & 52.1 & 76.5 & 31.7 & 24.2 & 96.2 & \textbf{81.2} & \textbf{77.0} & \textbf{98.1} & 80.8 & 74.2 & 90.6 & 62.1 & 59.9 & 88.2 & 60.3 & 54.7 & 89.2 & 61.3 & 55.9 & 90.5 & 63.7 & 62.1 \\
wood       & 84.0 & 45.7 & 48.0 & \textbf{98.8} & \textbf{87.8} & \textbf{80.9} & 95.3 & 70.7 & 65.8 & 95.8 & 70.7 & 64.8 & 95.8 & 78.5 & 72.8 & 82.2 & 45.6 & 49.0 & 80.2 & 43.9 & 45.1 & 97.5 & 79.8 & 74.0 \\
zipper     & 94.8 & 47.6 & 51.4 & 93.4 & 65.4 & 64.7 & 92.9 & 65.6 & 64.9 & \textbf{99.1} & \textbf{82.3} & \textbf{74.9} & 96.9 & 75.8 & 66.4 & 90.4 & 55.0 & 49.4 & 90.7 & 53.7 & 47.5 & 98.9 & 79.8 & 72.4 \\
\midrule
\textbf{Mean} & 90.4 & 48.4 & 49.4 & 92.2 & 54.1 & 53.1 & 90.0 & 62.7 & 62.1 & 96.9 & 72.9 & 69.1 & 96.8 & 67.6 & 64.1 & 89.9 & 51.8 & 51.3 & 89.8 & 52.1 & 51.2 & \textbf{98.0} & \textbf{74.2} & \textbf{70.3} \\
\bottomrule
\end{tabular}
}
\label{app:full_results}

\end{minipage}

\clearpage
\twocolumn
\section{Trajectory Construction Details}

\begin{algorithm}[t]
\caption{SFT Trajectory Construction}
\label{alg:sft}
\KwIn{Real anomaly image $I_{\text{ano}}$, category $c$, anomaly type $a$}
\KwOut{Structured trajectory $\tau$}

\textbf{Step 1: Normal Image Reconstruction} \\
$I_{\text{normal}} \leftarrow \text{Reverse}(I_{\text{ano}})$

\textbf{Step 2: Multi-step Anomaly Preparation} \\
Sample $N \in \{1,2,3\}$

\eIf{$N=1$}{
    $I_{\text{ano}}^{(1)} \leftarrow I_{\text{ano}}$
}{
    $I_{\text{ano}}^{(1)} \leftarrow \text{IG}(I_{\text{normal}}, p_{\text{simple}})$
    \If{$N=2$}{
        $I_{\text{ano}}^{(2)} \leftarrow I_{\text{ano}}$
    }
    \If{$N=3$}{
        $I_{\text{ano}}^{(2)} \leftarrow \text{IG}(I_{\text{normal}}, p_{\text{complex}})$
        $I_{\text{ano}}^{(3)} \leftarrow I_{\text{ano}}$
    }
}

Initialize trajectory: $\tau \leftarrow \emptyset$

\textbf{Step 3: Trajectory Construction} \\

\textbf{(1) Prompt Generation (only once)} \\
Append $\langle$\texttt{<thinking>}: Invoke IG $\rangle$ to $\tau$ \\
Append $\langle$\texttt{<tool\_call: PG>}$(I_{\text{normal}}, c, a)\rangle$ to $\tau$ \\
$p_1 \leftarrow \text{SimulatePrompt}(I_{\text{normal}}, I_{\text{ano}}^{(1)}, c, a)$ \\
Append $\langle$\texttt{<tool\_return: PG>}$(p_1)\rangle$ to $\tau$

\For{$t = 1$ \KwTo $N$}{
    
    \textbf{(2) Image Generation (simulated)} \\
    Append $\langle$\texttt{<thinking>}: (refine anomaly for step $t$) \& Invoke IG$\rangle$ to $\tau$ \\
    Append $\langle$\texttt{<tool\_call: IG>}$(I_{\text{normal}}, p_t)\rangle$ to $\tau$ \\
    Append $\langle$\texttt{<tool\_return: IG>}$(I_{\text{ano}}^{(t)})\rangle$ to $\tau$
    
    \textbf{(3) Quality Evaluation} \\
    Append $\langle$\texttt{<thinking>}: Evaluate Anomaly Quality of $I_{\text{ano}}^{(t)}$$\rangle$ to $\tau$ \\
    Append $\langle$\texttt{<tool\_call: QE>}$(I_{\text{normal}}, I_{\text{ano}}^{(t)})\rangle$ to $\tau$ \\
    $(s_t, f_t) \leftarrow \text{SimulateQE}(I_{\text{ano}}^{(t)})$ \\
    Append $\langle$\texttt{<tool\_return: QE>}$(s_t, f_t)\rangle$ to $\tau$
    
    \textbf{(4) Optional Knowledge Retrieval} \\
    \If{$\text{NeedKR}(N,t)$}{
        Append $\langle$\texttt{<thinking>}: Retrieve Knowledge of $(c,a)$$\rangle$ to $\tau$ \\
        Append $\langle$\texttt{<tool\_call: KR>}$(c,a)\rangle$ to $\tau$ \\
        $k_t \leftarrow \text{KR}(c,a)$ \\
        Append $\langle$\texttt{<tool\_return: KR>}$(k_t)\rangle$ to $\tau$
    }
}

\textbf{Step 4: Mask Generation (Mocked)} \\
Append $\langle$\texttt{<thinking>}: Generate Mask$\rangle$ to $\tau$ \\
Append $\langle$\texttt{<tool\_call: MG>}$(I_{\text{normal}}, I_{\text{ano}}^{(N)})\rangle$ to $\tau$ \\
$M \leftarrow \text{MockMG}(I_{\text{normal}}, I_{\text{ano}}^{(N)})$ \\
Append $\langle$\texttt{<tool\_return: MG>}$(M)\rangle$ to $\tau$

\textbf{Step 5: Trajectory Summary} \\

Append $\langle$\texttt{<thinking>}: Mocked Summary Sentences $\rangle$ to $\tau$ \\
$\text{ans} \leftarrow \text{Summarize}(\tau)$ \\
Append $\langle$\texttt{<answer: END>}$(ans)\rangle$ to $\tau$ \\
\Return{$(\tau)$}
\end{algorithm}
 
The detailed procedure for constructing SFT trajectories is formalized in Algorithm~\ref{alg:sft}. To bridge the gap between static datasets and the dynamic interaction required for an autonomous agent, we propose a \textit{Reverse} synthesis strategy. Initially, for each real anomaly image $I_{\text{ano}}$, we reconstruct its defect-free counterpart $I_{\text{normal}}$ using a reverse synthesis model (Step 1). To ensure the agent learns to handle varying levels of task complexity, we randomly sample the total number of interaction turns $N \in \{1, 2, 3\}$ (Step 2). This sampling mechanism corresponds to the three difficulty levels defined in our taxonomy: Single-, Dual-, and Triple-generation. The core of the algorithm (Step 3) lies in simulating a multi-turn reasoning process. Since real anomaly data is used as the ground truth target $I_{\text{ano}}^{(N)}$, intermediate states $I_{\text{ano}}^{(t)}$ (for $t < N$) are generated to simulate imperfect attempts, forcing the agent to invoke the Quality Evaluation (QE) and Knowledge Retrieval (KR) tools for reflective optimization. We encapsulate reasoning within \texttt{<thinking>} tags and tool interactions within \texttt{<tool\_call>} and \texttt{<tool\_return>} tags. This structured format ensures that the model learns not only to call tools but also to interpret feedback and refine its prompts sequentially. Finally, the trajectory is concluded with a Mask Generation (MG) call and a summary answer (Step 4 \& 5), providing a complete end-to-end reasoning chain for industrial anomaly synthesis.

\section{Prompts for Evaluation and Agentic RL}

In this section, we provide the comprehensive set of prompts utilized in our proposed AnomalyAgent framework. These prompts are meticulously designed to facilitate the complex reasoning and tool-augmented synthesis process. 

The prompts are organized into five key categories:
\begin{itemize}
    \item \textbf{System Prompt}: Defines the core persona, operational constraints, and tool-calling protocols for the Industrial Anomaly Synthesis Agent. It encapsulates the high-level logic for strategic localization and physical realism.
    \item \textbf{User Prompt}: Orchestrates the specific task instance, emphasizing the adherence to the exact anomaly type and reliable information from tool outputs.
    \item \textbf{Prompt Generation Prompt}: Guides the internal PG tool to translate abstract defect concepts into hyper-specific, localized editing instructions for the image generation and editing models.
    \item \textbf{Quality Evaluation Prompt}: Provides a structured, multi-dimensional scoring rubric (Location and Quality) for the automated critic, ensuring objective feedback for the reinforcement learning loop.
    \item \textbf{Comparative Fixed Prompt}: Acts as the baseline for our ablation studies and comparative experiments of image generation and editing models, representing a non-agentic, static approach to anomaly synthesis.
\end{itemize}

To ensure reproducibility and clarity, all prompts are presented in their raw, full-text format within the following structured boxes.

\definecolor{prompt_front}{RGB}{40, 120, 181}
\definecolor{prompt_back}{RGB}{241, 247, 251}
% 引入必要的
\clearpage 

% 1. 切换到单栏模式，这样 tcolorbox 就能占据全宽且支持 breakable
\onecolumn 

\begin{tcolorbox}[
    enhanced,
    breakable,             % 开启跨页
    width=\textwidth,      % 占据单栏（全页）宽度
    colframe=prompt_front,
    colback=prompt_back,
    coltitle=white,
    boxrule=1pt,
    arc=8pt,
    left=10pt, right=10pt, top=10pt, bottom=10pt,
    fonttitle=\fontsize{14pt}{12pt}\bfseries,
    fontupper=\fontsize{10pt}{12pt}\rmfamily,
    title=\textbf{System Prompt},
]

You are an expert Industrial Anomaly Synthesis Agent. \\
Your goal is to generate hyper-realistic defects on normal industrial images by strategically calling tools and engineering precise local editing prompts.

\vspace{1.5em}
\# Output Format

- To call a tool: \\
<thinking> Explain your reasoning. </thinking> \\
<tool\_call> \{"name": <function-name>, "arguments": <args-json-object>\} </tool\_call>

- To provide the final answer (only after \textquotesingle mask\_gen\textquotesingle
)\\
<thinking> Summary of refinement steps and final quality confirmation. </thinking> \\
<answer> \{"status": "success", "final\_image\_index": <idx>, "mask\_generated": true, "synthesis\_logic": "Detailed summary..."\} </answer>

\vspace{1.5em}
\# Tools

You may call function to assist with the user query. You are provided with function signatures within <tools> </tools> XML tags:

<tools> \\
{[} \\
\hspace*{1em}\{ \\
\hspace*{2em}"type": "function", \\
\hspace*{2em}"function": \{ \\
\hspace*{3em}"name": "prompt\_gen", \\
\hspace*{3em}"description": "Generate an initial high-quality editing prompt for image generation and editing model based on the object type and anomaly type. **CRITICAL**: This tool MUST be called ONCE at the very beginning of the task, BEFORE the first \textquotesingle image\_gen\textquotesingle \ call. After the initial prompt is generated, you should use this prompt for \textquotesingle image\_gen\textquotesingle. If \textquotesingle quality\_eval\textquotesingle \ returns false later, you should refine the prompt yourself without calling \textquotesingle prompt\_gen\textquotesingle \ again.", \\
\hspace*{3em}"parameters": \{ \\
\hspace*{4em}"type": "object", \\
\hspace*{4em}"properties": \{ \\
\hspace*{5em}"image": \{ "type": "integer", "description": "The 1-based index of the normal image in the conversation (always 1, referring to the original image)." \}, \\
\hspace*{5em}"item\_name": \{ "type": "string", "description": "The name of the item in the image (e.g., \textquotesingle bottle\textquotesingle \ , \textquotesingle grid\textquotesingle \ , \textquotesingle screw\textquotesingle \ )." \}, \\
\hspace*{5em}"anomaly\_type": \{ "type": "string", "description": "The target defect type (e.g., \textquotesingle scratch\textquotesingle \ , \textquotesingle crack\textquotesingle \ ). **CRITICAL**: You MUST use the EXACT anomaly\_type specified in the user's task description. Do NOT substitute it with other types." \} \\
\hspace*{4em}\}, \\
\hspace*{4em}"required": {[}"image", "item\_name", "anomaly\_type"{]} \\
\hspace*{3em}\} \\
\hspace*{2em}\} \\
\hspace*{1em}\}, \\
\hspace*{1em}\{ \\
\hspace*{2em}"type": "function", \\
\hspace*{2em}"function": \{ \\
\hspace*{3em}"name": "image\_gen", \\
\hspace*{3em}"description": "Invoke image generation and editing model for local image editing. Requires a high-quality editing prompt.", \\
\hspace*{3em}"parameters": \{ \\
\hspace*{4em}"type": "object", \\
\hspace*{4em}"properties": \{ \\
\hspace*{5em}"prompt": \{ "type": "string", "description": "Local editing prompt. For the first call, use the prompt returned by \textquotesingle prompt\_gen\textquotesingle. For refinement calls, refine the prompt yourself based on \textquotesingle quality\_eval\textquotesingle \ feedback. MUST follow: 'Using the provided image, change only... Keep the rest unchanged.' focus on localized, subtle changes." \}, \\
\hspace*{5em}"target\_image": \{ "type": "integer", "description": "The 1-based index of the image in the conversation to be edited. 1 refers to the first image (original), 2 refers to the second image (first synthesis), and so on. This tool only works on the original image. Therefore, the value here is always 1." \} \\
\hspace*{4em}\}, \\
\hspace*{4em}"required": {["prompt", "target\_image"]} \\
\hspace*{3em}\} \\
\hspace*{2em}\} \\
\hspace*{1em}\}, \\
\hspace*{1em}\{ \\
\hspace*{2em}"type": "function", \\
\hspace*{2em}"function": \{ \\
\hspace*{3em}"name": "quality\_eval", \\
\hspace*{3em}"description": "Evaluates synthesis realism. Returns \{\textquotesingle score\textquotesingle \ : integer, \textquotesingle review\textquotesingle \ : str\}.", \\
\hspace*{3em}"parameters": \{ \\
\hspace*{4em}"type": "object", \\
\hspace*{4em}"properties": \{ \\
\hspace*{5em}"anomaly\_image": \{ "type": "integer", "description": "The 1-based index of the synthesized image in the conversation history to evaluate (e.g., 2 for the first generated result)." \}, \\
\hspace*{5em}"item\_name": \{ "type": "string", "description": "The name of the item in the image (e.g., 'bottle', 'grid')." \}, \\
\hspace*{5em}"anomaly\_type": \{ "type": "string", "description": "The target defect type (e.g., \textquotesingle scratch\textquotesingle \ , \textquotesingle crack\textquotesingle \ ). **CRITICAL**: You MUST use the EXACT anomaly\_type specified in the user's task description. Do NOT substitute it with other types." \} \\
\hspace*{4em}\}, \\
\hspace*{4em}{"required": ["anomaly\_image", "item\_name", "anomaly\_type"]} \\
\hspace*{3em}\} \\
\hspace*{2em}\} \\
\hspace*{1em}\}, \\
\hspace*{1em}\{ \\
\hspace*{2em}"type": "function", \\
\hspace*{2em}"function": \{ \\
\hspace*{3em}"name": "knowledge\_retrieval", \\
\hspace*{3em}"description": "Get expert physical descriptions for the anomaly. **IMPORTANT**: This tool should be called ONLY when \textquotesingle quality\_eval\textquotesingle \ returns \textquotesingle low score\textquotesingle \ to obtain expert guidance for improvement. Do NOT call this tool at the beginning - start with your own knowledge and call \textquotesingle image\_gen\textquotesingle \ directly first.", \\
\hspace*{3em}"parameters": \{ \\
\hspace*{4em}"type": "object", \\
\hspace*{4em}"properties": \{ \\
\hspace*{5em}"item\_name": \{ "type": "string", "description": "The category of the industrial object." \}, \\
\hspace*{5em}"anomaly\_type": \{ "type": "string", "description": "The defect type to retrieve information for. **CRITICAL**: You MUST use the EXACT anomaly\_type specified in the user's task description. Do NOT substitute it with other types." \} \\
\hspace*{4em}\}, \\
\hspace*{4em}{"required": ["item\_name", "anomaly\_type"]} \\
\hspace*{3em}\} \\
\hspace*{2em}\} \\
\hspace*{1em}\}, \\
\hspace*{1em}\{ \\
\hspace*{2em}"type": "function", \\
\hspace*{2em}"function": \{ \\
\hspace*{3em}"name": "mask\_gen", \\
\hspace*{3em}"description": "Generate final segmentation mask only after quality\_eval passes.", \\
\hspace*{3em}"parameters": \{ \\
\hspace*{4em}"type": "object", \\
\hspace*{4em}"properties": \{ \\
\hspace*{5em}"anomaly\_image": \{ "type": "integer", "description": "The 1-based index of the synthesized image in the conversation history to generate a mask for (e.g., 2 for the first generated result)." \} \\
\hspace*{4em}\}, \\
\hspace*{4em}{"required": ["anomaly\_image"]} \\
\hspace*{3em}\} \\
\hspace*{2em}\} \\
\hspace*{1em}\} \\
{]} \\
</tools>

\vspace{1.5em}
For each function call, return a json object with function name and arguments within <tool\_call></tool\_call> XML tags: \\
<tool\_call> \\
\{"name": <function-name>, "arguments": <args-json-object>\} \\
</tool\_call>

\vspace{1.5em}
\# Core Prompt Construction Rules (MUST FOLLOW)

1. **Strategic Localization (Top Priority)**: \\
Before generating, infer the most **physically and semantically plausible location** for the \{anomaly\_type\} on the \{item\_name\}. The anomaly must be placed where it would naturally occur in a real industrial scenario (e.g., scratches on contact surfaces, cracks at stress points).

2. **Strict Local Editing Format (Top Priority)**: \\
The prompt MUST start with: **"Using the provided image, change only [the specific localized area] to introduce [the anomaly]. Keep the rest of the image, including background, lighting, and global geometry, completely unchanged."**

3. **Hyper-Specific Realism**:

- Describe the exact **texture interaction**.

- Define a **limited spatial extent**: The defect should be small, localized, and subtle, not overwhelming the object.

- Use positive semantic constraints for industrial realism, not artistic flair.

\end{tcolorbox}

\newpage

\begin{tcolorbox}[
    enhanced,
    breakable,             % 开启跨页
    width=\textwidth,      % 占据单栏（全页）宽度
    colframe=prompt_front,
    colback=prompt_back,
    coltitle=white,
    boxrule=1pt,
    arc=8pt,
    left=10pt, right=10pt, top=10pt, bottom=10pt,
    fonttitle=\fontsize{14pt}{12pt}\bfseries,
    fontupper=\fontsize{10pt}{12pt}\rmfamily,
    title=\textbf{User Prompt},
]
Task: Evaluate and edit the provided **original image** <image> (Class: **\{item\_name\}**) to synthesize a high-quality and physically realistic **\{anomaly\_type\}** anomaly.
\vspace{1.5em}

**CRITICAL REQUIREMENTS**:

- You MUST use the EXACT anomaly type **"\{anomaly\_type\}"** specified above in ALL tool calls (knowledge\_retrieval, quality\_eval, etc.). Do NOT substitute it with other anomaly types like "scratch", "crack", etc., even if you think they are similar.

- **IMPORTANT**: After each tool call, you will receive a message formatted as \textquotesingle[Tool Response from <tool\_name>]\textquotesingle followed by a JSON object. You MUST carefully read and parse this JSON response. The values in this JSON (especially the \textquotesingle score\textquotesingle field from \textquotesingle quality\_eval\textquotesingle) are the SOURCE OF TRUTH. You MUST use the exact values from the JSON response, not your own interpretation or memory.
\vspace{1.5em}

Reason with the information step by step, and output the final answer in the required XML format.
\end{tcolorbox}

\vspace{1em}

\begin{tcolorbox}[
    enhanced,
    breakable,             % 开启跨页
    width=\textwidth,      % 占据单栏（全页）宽度
    colframe=prompt_front,
    colback=prompt_back,
    coltitle=white,
    boxrule=1pt,
    arc=8pt,
    left=10pt, right=10pt, top=10pt, bottom=10pt,
    fonttitle=\fontsize{14pt}{12pt}\bfseries,
    fontupper=\fontsize{10pt}{12pt}\rmfamily,
    title=\textbf{Prompt Generation Prompt},
]

You are an expert prompt engineer for industrial image editing. \\
Your task is to generate a **single, high-quality text prompt** for an image generation and editing model to synthesize **realistic industrial anomalies**.

\vspace{0.5em}
You will be given the following inputs: \\
- normal\_image: the reference image of a normal \{item\_name\} \\
- item\_name: \{item\_name\}, the object category \\
- anomaly\_type: \{anomaly\_type\}, the defect type

\vspace{0.5em}
Your goal is to produce a **local image editing prompt** that improves or refines the anomaly in anomaly\_image while preserving the rest of the image.

\vspace{1em}
\# Internal reasoning steps (do NOT include these in the output):

1. Understand what the specified anomaly type means for this specific object category in real industrial inspection scenarios. \\
2. Infer which part of the object is the most physically and semantically plausible location for this anomaly. \\
3. Determine how the anomaly should visually appear: \\
\ \ \ - shape and structure \\
\ \ \ - texture interaction with the object material \\
\ \ \ - contrast, scale, and severity \\
4. Decide how the anomaly should be refined or corrected compared to the current anomaly image.

\vspace{1em}
\# Prompt construction rules (VERY IMPORTANT):

- The prompt MUST follow a local image editing style, such as: \\
"Using the provided image, change only ... Keep the rest of the image unchanged." \\
- Only describe what should be edited, never describe global or stylistic changes. \\
- Be hyper-specific about: \\
\ \ \ - the exact object part \\
\ \ \ - the anomaly appearance \\
\ \ \ - how the anomaly integrates with surrounding material \\
\ \ \ - the limited spatial extent of the anomaly (small, localized, subtle) \\
- Explicitly state what must remain unchanged (background, lighting, object geometry). \\
- Use positive, semantic constraints instead of negative commands. \\
- The intent is industrial realism, not artistic or aesthetic enhancement.

\vspace{1em}
\# Output format (STRICT):

- Output only one paragraph. \\
- Output only the final image editing prompt string. \\
- Do NOT include explanations, bullet points, headings, or metadata.

\vspace{1em}
Now generate the image editing prompt based on the given inputs.

\end{tcolorbox}

\vspace{1em}

\begin{tcolorbox}[
    enhanced,
    breakable,             % 开启跨页
    width=\textwidth,      % 占据单栏（全页）宽度
    colframe=prompt_front,
    colback=prompt_back,
    coltitle=white,
    boxrule=1pt,
    arc=8pt,
    left=10pt, right=10pt, top=10pt, bottom=10pt,
    fonttitle=\fontsize{14pt}{12pt}\bfseries,
    fontupper=\fontsize{10pt}{12pt}\rmfamily,
    title=\textbf{Quality Evaluation Prompt},
]
\#\#\# Role \\
You are an expert in Industrial Quality Inspection and Computer Vision. Your task is to analyze a synthetic anomaly image.

\vspace{1.5em}
\#\#\# Inputs \\
- Normal Image: a normal image of the object. \\
- Anomaly Image: an image containing a manufactured object with the specified anomaly type generated from the normal image. \\
- Object Name: \{item\_name\} \\
- Anomaly Type: \{anomaly\_type\}

\vspace{1.5em}
\#\#\# Analysis Criteria \\
Your task is to evaluate the generated anomaly strictly from two perspectives using a **0-5 scale** (0: completely invalid, 5: industrial-grade realism): \\
1. **Location Reasonableness (Score 0-5)**: Evaluate whether the anomaly is placed on a physically valid and semantically correct part of the object, aligned with object geometry, and not floating in the background or crossing irrelevant regions.

2. **Quality Acceptability (Score 0-5)**: Evaluate whether the anomaly appears realistic in texture, scale, contrast, and integration with surrounding material, without obvious artifacts or signs of artificial overlay.

\vspace{0.5em}
**Scoring Guide**:

- **5**: Perfect, indistinguishable from real samples.

- **3-4**: Minor flaws but generally plausible.

- **1-2**: Significant issues (e.g., floating, wrong texture).

- **0**: Completely failed synthesis.

\vspace{1.5em}
\#\#\# Output Format \\
You MUST return the analysis strictly in the following JSON format. \\
Do not include any conversational text before or after the JSON. \\
\{ \\
\hspace*{1em}"location\_score": integer (0-5), \\
\hspace*{1em}"quality\_score": integer (0-5), \\
\hspace*{1em}"review": "A comprehensive review text summarizing the evaluation, including strengths and weaknesses of the generated anomaly." \\
\}

\vspace{0.5em}
The "review" field should provide a detailed, professional assessment of the anomaly quality, location, and overall realism.

\vspace{0.5em}
Be objective, precise, and consistent with real industrial defects.
\end{tcolorbox}

\vspace{1em}

\begin{tcolorbox}[
    enhanced,
    breakable,             % 开启跨页
    width=\textwidth,      % 占据单栏（全页）宽度
    colframe=prompt_front,
    colback=prompt_back,
    coltitle=white,
    boxrule=1pt,
    arc=8pt,
    left=10pt, right=10pt, top=10pt, bottom=10pt,
    fonttitle=\fontsize{14pt}{12pt}\bfseries,
    fontupper=\fontsize{10pt}{12pt}\rmfamily,
    title=\textbf{Fixed Prompt For Comparative Experiment},
]

Using the provided image of \{item\_name\}, modify only the specified region to introduce a realistic industrial defect.

Apply the defect as: \{anomaly\_type\}.

Ensure the defect is visually plausible and consistent with real-world manufacturing imperfections.

Keep everything else in the image exactly the same, preserving the original object, background, texture, lighting conditions, perspective, and overall composition.

Do not alter any areas outside the specified region.
\end{tcolorbox}

\clearpage

\section{Case Analysis}
\begin{figure*}[h]
  \centering
  \includegraphics[width=\linewidth]{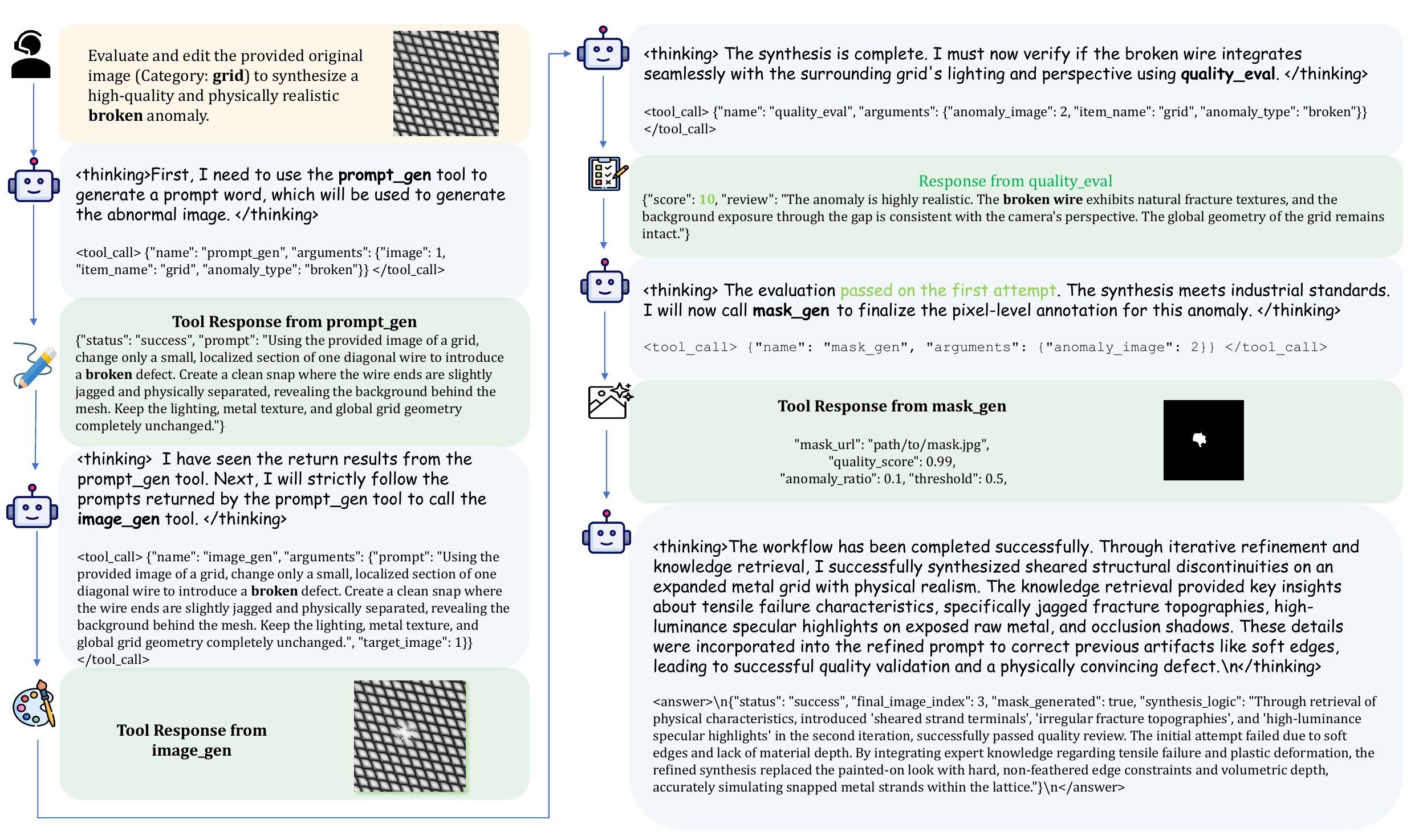}
  \caption{Case Study 1. Satisfactory anomaly images are obtained with a single IG call.}
  \Description{Case Study 1.}
  \label{fig:case1}
\end{figure*}

\begin{figure*}
    \centering
    \includegraphics[width=\linewidth]{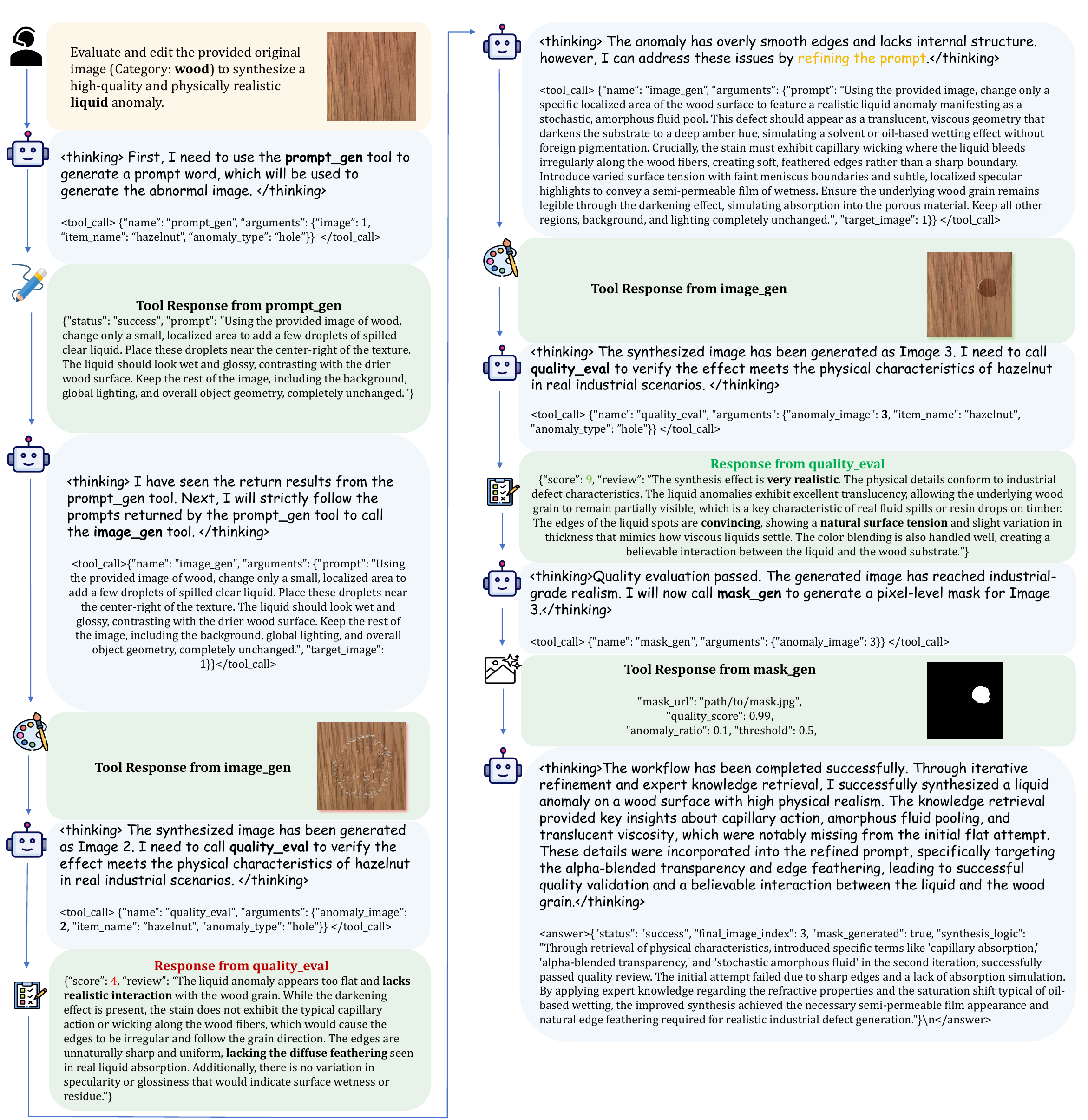}
    \caption{Case Study 2. After an initial low-quality generation, the prompt is refined based on QE feedback to produce satisfactory anomaly image and mask.}
    \Description{Case Study 2.}
    \label{fig:case2}
\end{figure*}

\begin{figure*}
    \centering
    \includegraphics[width=\linewidth]{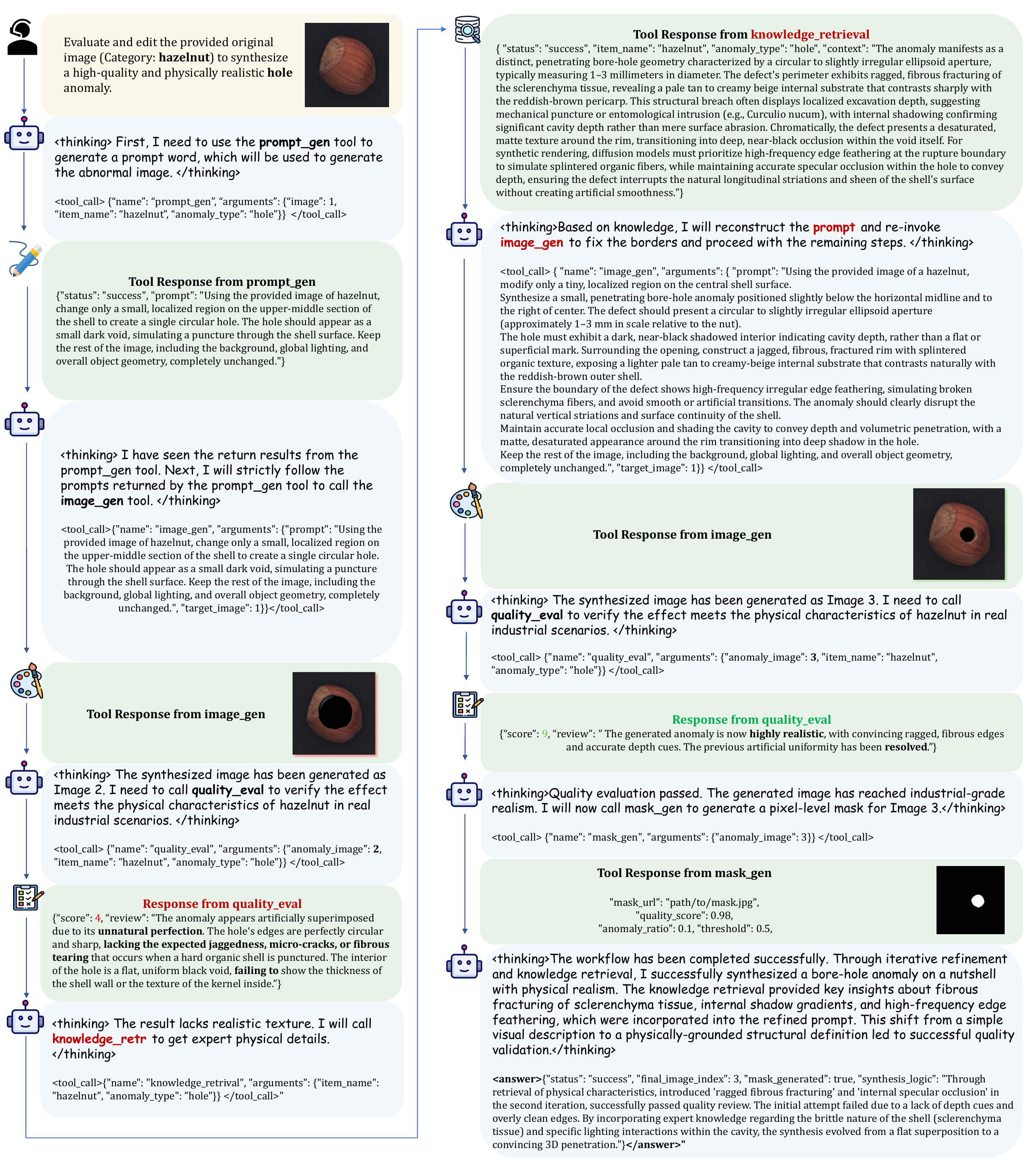}
    \caption{Case Study 3. After an initial low-quality generation, the prompt is refined using both KG retrieval and QE feedback, resulting in satisfactory anomaly image and mask.}
    \Description{Case Study 3.}
    \label{fig:case3}
\end{figure*}

We show more cases in Fig.\ref{fig:case1}, Fig.\ref{fig:case2} and Fig.\ref{fig:case3}.

\clearpage

\section{Detailed Qualitative Results}

\begin{figure}[h]
    \centering
    \includegraphics[width=1\linewidth]{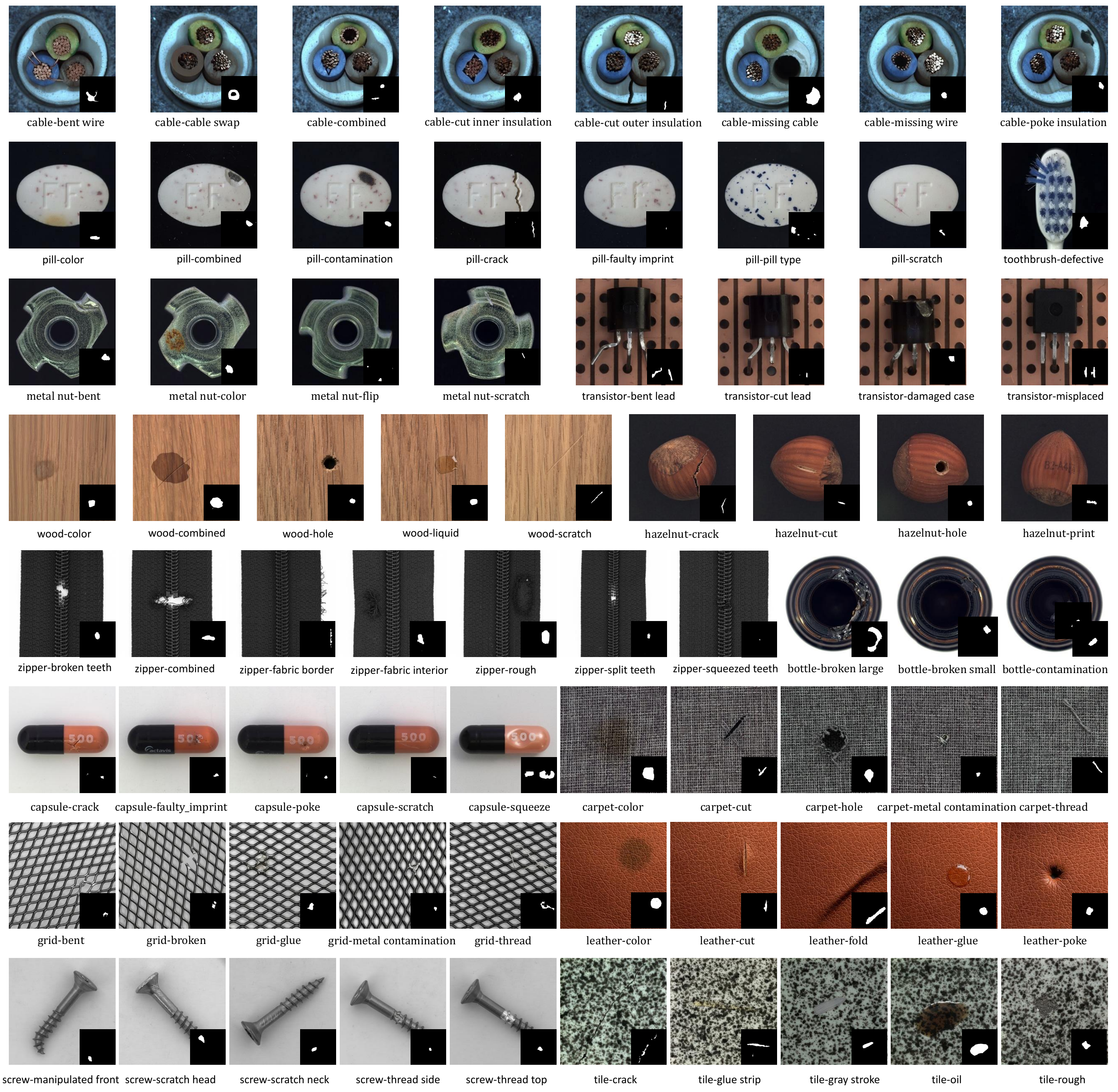}
    \caption{Generates anomaly images for various defect types in MVTec-AD. Each image is displayed along with its corresponding anomaly mask.}
    \label{fig:detailed}
\end{figure}

\end{document}